%% file: document4.tex
 \documentclass[journal]{IEEEtran}

\usepackage{cs}
\usepackage{mathsymb}
\usepackage{algorithm}
\usepackage{algorithmic}
\usepackage{graphicx}
\usepackage{url}
\usepackage{cases}
\usepackage{subfigure}
\graphicspath{{./}{./Fig/}{./Figs/}}

\renewcommand{\algorithmicrequire}{\textbf{Input:}}
\renewcommand{\algorithmicensure}{\textbf{Output:}}

\usepackage{rotating}

\def\bW{{\mathbf W} }
\def\bZ{{\mathbf Z} }
\def\bY{{\mathbf Y} }
\def\bX{{\mathbf X} }
\def\bS{{\mathbf S} }
\def\bx{{\mathbf x} }
\def\bQ{{\mathbf Q} }
\def\bH{{\mathbf H} }
\def\bP{{\mathbf p} }
\def\bA{{\bar{\mathbf{A}}} }
\def\wlda{SD-WLDA}

\def\psd{p.s.d\onedot}

\newtheorem{theorem}{Theorem}[section]

\newtheorem{proposition}{Proposition}[section]

\begin{document}

\title{Worst-Case Linear Discriminant Analysis as Scalable Semidefinite Feasibility Problems}

\author{Hui Li,
        Chunhua Shen,
        Anton van den Hengel, Qinfeng Shi
\thanks{The authors are with School of Computer Science, The
University of Adelaide, Australia.
C. Shen and A. van den Hengel are also with Australian Centre for Robotic Vision.
Correspondence should be addressed to C.
Shen (chunhua.shen@adelaide.edu.au).}
}

\markboth{Manuscript}%
{Li \MakeLowercase{\textit{et al.}}:
Worst-Case Linear Discriminant Analysis as Scalable Semidefinite Feasibility Problems}

\maketitle

\input{document4_intro.tex}

\input{document4_main.tex}

\input{document4_exp.tex}
\input{document4_conclusion.tex}

\bibliographystyle{IEEEtran}
\bibliography{IEEEabrv,mybibfile}

\clearpage

\input{Appendix.tex}

\end{document}

%% file: document4_intro.tex
\begin{abstract}

In this paper, we propose an efficient semidefinite programming (SDP) approach
to worst-case linear discriminant analysis (WLDA).
Compared with the traditional LDA,
WLDA considers the dimensionality reduction problem from the worst-case viewpoint,
which is in general more robust for classification.
However, the original problem of WLDA is non-convex and difficult to optimize.
In this paper, we reformulate the optimization problem of WLDA into a sequence of semidefinite feasibility problems.
To efficiently solve the semidefinite feasibility problems, we design a new scalable optimization
method with quasi-Newton methods and eigen-decomposition being the core components.
The proposed method is orders of magnitude faster than standard
interior-point based SDP solvers.

Experiments on a variety of classification problems demonstrate that our approach achieves
better performance than standard LDA.
Our method is also much faster and more scalable than
standard interior-point SDP solvers based WLDA.
The computational complexity for an SDP with $m$ constraints and matrices of size $d$ by $d$
is roughly reduced from $\mathcal{O}(m^3+md^3+m^2d^2)$ to $\mathcal{O}(d^3)$ ($m>d$ in our case).
\end{abstract}


\section{Introduction}
\label{sec:intro}

Dimensionality reduction is a critical problem in machine learning, pattern recognition and computer vision,
which for linear case learns a transformation matrix $\bW\!\in \Real^{d \times r}$ $(r \leq d)$
to project the input data $\bx\!\in\!\Real^d$
to a lower-dimensional space $\by = \bW^\T \bx \in \Real^r$
 such that the important structure or geometry of input data is preserved.
It can help us to eliminate the inherent noise of
data, and improve the classification performance. It can also decrease the
computational complexities of subsequent machine learning tasks.
There are two classical dimensionality reduction methods used widely in many applications,
principal component analysis (PCA) and linear discriminant analysis
(LDA). PCA is
an unsupervised linear dimensionality reduction method, which seeks a
subspace of the data that have the maximum variance and subsequently
projects the input data onto it. PCA may not give good classification performance due to its unsupervised nature.
LDA is in supervised fashion, which aims to maximize the {\em average} distance between each two class means
and minimize the {\em average} distance between each two samples within the same class. However, it has some limitations:
1) For $c$-class data, the target dimension of the projected subspace is restricted to be
at most $(c-1)$. In this sense, LDA is suboptimal and may cause so called {\em class separation} problem
that LDA tends to merge classes which are close in the original space;
2) It assumes that conditional probability density functions are normally distributed, which does not hold in many cases;
3) The scatter matrices are required to be nonsingular.

There are a large number of extension works to LDA and
PCA~\cite{shental2002adjustment,weinberger2006distance,chen2000new,ye2005characterization,Zhang2010NIPS,Gao2012enhanced,Tao2009geometric, Bian2008harmonic, Bian2011, Bian2012asymptotic}.
Among these methods, our focus in this paper is on the discrimination criterion of
worst-case linear discriminant analysis (WLDA), which was proposed by Zhang \etal~\cite{Zhang2010NIPS}.
Instead of using {\em average} between-class and within-class distances as LDA,
WLDA considers scatter measures from the {\em worst-case} view,
which arguably is more suitable for applications like classification.
Specifically, WLDA tries to
maximize the minimum of pairwise distances between class means, and
minimize the maximum of within-class pairwise distances over all classes.
Due to the complex formulation of its criterion, the problem of WLDA is difficult to optimize.

In~\cite{Zhang2010NIPS}, the problem of WLDA was firstly relaxed to a metric learning
problem on $\bZ\!=\!\bW \bW^{\T}$, which can be solved by a sequence of SDP optimization
procedures, where SDP problems are solved by standard interior-point methods (We denote it as Zhang~\etal~(SDP)).
However,
standard interior-point SDP solvers
scale poorly to problems with high dimensionality,
as the computational complexity is $\mathcal{O}(m^3+md^3+m^2d^2)$,  where $d$ is the problem size, and $m$ is the number of constraints in SDP.
An alternative optimization procedure was then proposed in~\cite{Zhang2010NIPS}
for large-scale WLDA problems, in which one column of the
transformation matrix was found at each iteration while fixing the other directions. We denote this method as Zhang~\etal~(SOCP) since it was reformulated as a series of second-order cone programming (SOCP) problems lastly.
Typically, this greedy method does not guarantee to find a globally
optimal solution.

In this paper we propose a fast SDP approach to solve WLDA problem.
The problem is converted to a sequence of SDP feasibility problems
using bisection search strategy, which can find a globally optimal
solution to the relaxed problem.
More importantly, we adopt a novel approach to solve SDP
feasibility problem at each iteration.
Motivated by~\cite{Shen2011CVPR}, a Frobenius-norm regularized SDP
formulation is used, and its Lagrangian dual can be solved effectively
by quasi-Newton methods.
The computational complexity of this optimization method is dominated
by eigen-decomposition at each iteration, which is $\mathcal{O}(d^3)$. The proposed method is denoted as \wlda. The main contributions of this work are:
1) By introducing an auxiliary variable, the original WLDA problem is
reformulated and can be solved via a sequence of convex feasibility
problems, by which the global optimum can be obtained for the relaxed
metric learning problem.
2) By virtue of the use of Frobenius norm regularization,
the optimization problem can be addressed by solving its Lagrange dual,
where first-order methods such as quasi-Newton can be used.
This approach is much faster than solving the corresponding primal problem using
standard interior-point methods, and can be applied to large-scale problems.
Next, we briefly review some relevant work.

\textit{Dimensionality reduction}
In order to overcome the drawbacks of LDA and improve the accuracy in classification,
many extensions have been proposed, such as
relevant component analysis (RCA)~\cite{shental2002adjustment},
neighborhood component analysis (NCA)~\cite{weinberger2006distance},
null space LDA (NLDA)~\cite{chen2000new},
orthogonal LDA (OLDA)~\cite{ye2005characterization}, Enhanced fisher discriminant criterion (EFDC)~\cite{Gao2012enhanced}, Geometric mean-based subspace selection (GMSS)~\cite{Tao2009geometric},
Harmonic mean-based subspace selection (HMSS)~\cite{Bian2008harmonic}, and Max-min distance analysis (MMDA)~\cite{Bian2011}.
Assuming dimensions with large within-class covariance are not
relevant to subsequent classification tasks,
RCA~\cite{shental2002adjustment} assigns
large weights to ``relevant dimensions'' and small weights to
``irrelevant dimensions'', where the relevance is estimated using
equivalence constraints.
NCA~\cite{weinberger2006distance}
learns the transformation matrix $\bW$ directly by minimizing the expected
leave-one-out classification error of $k$-nearest neighbours on the transformed space.
Because the objective function to be optimized is not convex, NCA
tends to converge to a local optimum.
NLDA~\cite{chen2000new},
OLDA~\cite{ye2005characterization} and EFDC~\cite{Gao2012enhanced}  were proposed to
address the problem that standard LDA fails when scatter
matrices are singular.
NLDA maximizes the between-class distance in the null
space of the within-class scatter matrix,
while OLDA calculates a set of orthogonal discriminant vectors by diagonalizing the scatter matrices simultaneously.
The resulting transformation matrices are both orthogonal for NLDA and
OLDA, and they are equivalent to each other under a mild condition~\cite{Ye2006}. EFDC incorporates the intra-class variation into the Fisher discriminant criterion, so that data from the same class can be mapped to a subspace where both the intraclass compactness and intraclass variation are well preserved. In this way, this method is robust to the intraclass variation and results in a good generalization capability.
To avoid the class separation problem of LDA, Tao \etal~\cite{Tao2007DM} proposed a general averaged divergence analysis (GADA) framework, which presented a general mean function in place of the arithmetic mean used in LDA. By choosing different mean functions, several subspace selection algorithms have been developed.
GMSS~\cite{Tao2009geometric} investigates the effectiveness of the geometric mean-based subspace selection, which maximizes the geometric mean of Kullback-Leibler (KL) divergences between different class pairs. HMSS~\cite{Bian2008harmonic} maximizes the harmonic mean of the symmetric KL divergences between all class pairs. They adaptively give large weights to class pairs that are close to each other, and result in better class separation performance than LDA. Instead of assigning weights to class pairs, MMDA~\cite{Bian2011} directly maximizes the minimum pairwise distance of all class pairs in the low-dimensional subspace, which guarantees the separation of all class pairs. However, MMDA does not take into account of the within-class pairwise distances over all classes.
Recently, Bian \etal~\cite{Bian2012asymptotic} presented an asymptotic generalization analysis of LDA,  which enriched the existing theory of LDA further. They showed that the generalization ability of LDA is mainly determined by the ratio of dimensionality to training sample size, where both feature dimensionality and training data size can be proportionally large.

Many dimensionality reduction algorithm such as PCA and LDA can be formulated into a trace ratio optimization problem
\cite{jia2009revisited}.
Guo \etal~\cite{Guo2003generalized} presented
a generalized Fisher discriminant criterion, which is essentially a trace ratio.
They proposed a heuristic bisection way, which was proven to converge to the precise solution.
Wang \etal~\cite{wang2007trace} tackled the trace ratio problem
directly by an efficient iterative procedure, where a trace difference problem
was solved via the eigen-decomposition method in each step.
Shen \etal~\cite{shen2010geometric} provided a geometric revisit to the trace ratio problem in the framework of optimization on the Grassmann manifold. Different from~\cite{wang2007trace}, they proposed another efficient algorithm, which employed only one step of the parallel Rayleigh quotient iteration at each iteration. Kokiopoulou \etal~\cite{Kokiopoulou} also treated the dimensionality reduction problem as trace optimization problems, and gave an overview of the eigenvalue problems encountered in dimensionality reduction area. They made a comparition between nonlinear and linear methods for dimensionality reduction, including Locally Linear Embedding (LLE), Laplacean Eigenmaps, PCA, Locality Preserving Projections (LPP), LDA, \etc, and showed that all the eigenvalue problems in explicit linear projections can be regarded as projected analogues of the so-called nonlinear projections.

Different from the aforementioned methods, WLDA considers the dimensionality reduction problem from a worst-case viewpoint. It maximizes the worst-case between-class scatter matrix and minimizes the worst-case within-class scatter matrix simultaneously, which can lead to more robust classification performance. The inner maximization and minimization over discrete variables make it different from the general trace ratio problem, and difficult to solve. The method of solving the general trace ratio problem cannot be extended here directly. Furthermore, different from the iterative algorithm for trace ratio optimization problem~\cite{wang2007trace}, we formulate the WLDA problem as a sequence of SDP problems, and propose an efficient SDP solving method. The eigen-decomposition we used is to solve the Lagrange dual gradient, which differs from that employed in solving the trace ratio optimization problem.

\textit{Solving large-scale SDP problems} Instead of learning the transformation matrix $\bW$,
quadratic Mahalanobis distance metric learning methods (which are highly related to dimensionality reduction methods)
optimize over $\bZ\!=\!\bW \bW^{\T}$,
in order to obtain a convex relaxation.
The transformation matrix $\bW$ can be recovered from the eigen-decomposition of $\bZ$.
Because $\bZ$ is positive semidefinite (\psd) by definition, quadratic Mahalanobis metric learning methods optimizing on $\bZ$ usually need to solve an SDP problem.

Xing \etal~\cite{xing2003distance} formulated metric learning as a convex (SDP) optimization problem,
and a globally optimal solution can be obtained.
Weinberger \etal~\cite{Weinberger2009} presented a distance metric learning method,
which optimizes a Mahalanobis metric such that the $k$-nearest neighbours always belong to the same class while samples from different classes are separated by a large margin.
In terms of SDP solver, they proposed an alternate projection method, where the learned metric $\bZ$ is projected back onto the \psd cone by eigen-decomposition at each iteration.
MMDA~\cite{Bian2011} was solved approximately by a sequence of SDP problems using standard interior-point methods.
Shen \etal~\cite{Shen2008PR} proposed a novel SDP based method for directly
solving trace quotient problems for dimensionality reduction.
With this method, globally-optimal solutions can be obtained for trace quotient problems.

As we can see, many aforementioned methods used standard interior-point SDP solvers, which are unfortunately
computationally expensive (computational complexity is $\mathcal{O}(m^3+md^3+m^2d^2)$) and scale poorly to large-scale problems.
Thus an efficient SDP optimization approach is critical for large-scale metric learning problems.

There are many recent work to address large-scale SDP problems arising from distance metric learning and other computer vision tasks.
Shen \etal~\cite{Shen2011CVPR} proposed a fast SDP approach for solving Mahalanobis metric learning problem.
They introduced a Frobenius-norm regularization in the objective function of SDP problems,
which leads to a much simpler Lagrangian dual problem: the objective function is continuously differentiable and \psd constraints in the dual can be eliminated.
L-BFGS-B was used to solve the dual, where a partial eigen-decomposition needed to be calculated at each iteration.
Wang \etal~\cite{wang2013fast} also employed a similar dual approach to solve binary quadratic problems for computer vision tasks, such as image segmentation, co-segmentation, image registration.
SDP optimization method in~\cite{Shen2011CVPR,wang2013fast} can be seen as an extension of the works in~\cite{boyd2005least,malick2004dual},
which considered semidefinite least-squares problems.
The key motivation of~\cite{boyd2005least,malick2004dual} is that the objective function of the corresponding dual problem is continuously differentiable but not twice differentiable,
therefore first-order methods can be applied.
Malick~\cite{malick2004dual} and Boyd and Xiao~\cite{boyd2005least} proposed to use quasi-Newton methods and projected gradient methods respectively,
to solve the Lagrangian dual of semidefinite least-squares problems.
Semismooth Newton-CG methods~\cite{zhao2010newton} and smoothing Newton methods~\cite{gao2009calibrating} are also exploited for semidefinite least-squares problems,
which require much less number of iterations at the cost of higher computational complexity per iteration (full eigen-decomposition plus conjugate gradient).

Alternatively, stochastic (sub)gradient descent (SGD) methods~\cite{nemirovski2009robust} were also employed to solve SDP problems.
Combining with alternating direction methods~\cite{wen2010alternating,ouyang2013stochastic}, SGD can be used for SDP problems with inequality and equality constraints.
The computational bottleneck of typical SGD is the projection of one infeasible point onto the \psd cone at each iteration,
which leads to the eigen-decomposition of a $d \times d$ matrix.
A number of methods have been proposed to speed up the projection operation at each iteration.
Chen \etal~\cite{chen2014efficient} proposed a low-rank SGD method,
in which rank-$k$ stochastic gradient is constructed and then
the projection operation is simplified to compute at most $k$ eigenpairs.
In the works of~\cite{shalev2004online,davis2007information,jain2008online,shen2012positive},
the distance metric is updated by rank-one matrices iteratively, and no eigen-decomposition or only one leading eigenpair is required.
Note that SGD methods usually need more iterations to converge than the dual approaches based on quasi-Newton methods~\cite{Shen2011CVPR}.

The most related work to ours may be Shen \etal's~\cite{Shen2011CVPR}.
We use similar SDP optimization technique as that in~\cite{Shen2011CVPR}.
However, SDP feasibility problems are considered in our paper
while the work in~\cite{Shen2011CVPR} focuses on standard SDP problems with linear objective functions.

\textbf{Notation}  We use a bold lower-case letter $\bx$ to denote a column vector, and a bold capital letter $\bX$ to denote a matrix. $\bX^\T$ is the transposition of $\bX$. $\Real ^{m \times n}$ indicates the set of $m \times n$ matrices. $\eyes_n$ represents an $n \times n$ identity matrix.
$\bX \succcurlyeq \bY$ indicates that the matrix $\bX - \bY$ is positive semidefinite.
$\langle \cdot,\cdot \rangle$ denotes the inner product of two matrices or vectors.
$\trace(\cdot)$ indicates the trace of a matrix. $\|\cdot\|_F$  is the Frobenius norm of a matrix.
$\diag(\cdot)$ returns a diagonal matrix with the input elements on its main diagonal.
Suppose that the eigen-decomposition of
a symmetric matrix $\bX \in \Real^{n \times n}$ is $\bX = \mathbf{U} \diag(\lambda_1, \lambda_2, \dots, \lambda_n) \mathbf{U}^\T$, where $\mathbf{U}$ is
the orthonormal matrix of eigenvectors of $\bX$, and $\lambda_1, \dots, \lambda_n$ are the corresponding eigenvalues,
we define the positive and negative parts
of $\bX$ respectively as
\begin{align}
(\bX)_+ &= \mathbf{U} \diag(\max(\lambda_1,0), \dots \max(\lambda_n,0))\mathbf{U}^\T, \\
(\bX)_- &= \mathbf{U} \diag(\min(\lambda_1,0), \dots \min(\lambda_n,0))\mathbf{U}^\T
\end{align}
It is clear that  $\bX = (\bX)_+ + (\bX)_-$ holds.

%% file: document4_main.tex
\section{Worst-Case Linear Discriminant Analysis}
\label{sec:WLDA}

We briefly review WLDA problem proposed by~\cite{Zhang2010NIPS} firstly. Given a training set of $n$ samples $\mathcal{D} = \{ \bx_1, \dots, \bx_n \}$ ($\bx_n \in \Real^d$),
which consists of $c \geq 2$ classes $\Pi_i, i = 1, \dots, c$,
where class $\Pi_i$ contains $n_i$ samples. As we mentioned before, the aim of linear dimensionality reduction
is to find a transformation matrix $\bW \in  \Real^{d \times r}$ with $r \leq d$.

We define the within-class scatter matrix of the $k$th class $\Pi_k$ as
	\begin{equation}
	\bS_k = \frac{1}{n_k} \sum_{\bx_i \in \Pi_k} (\bx_i - \overline{\mathbf{m}}_k)
	 (\bx_i - \overline{\mathbf{m}}_k)^\T,
	\label{equ:covariance}
	\end{equation}
which is also the covariance matrix for the $k$th class, and $ \overline{\mathbf{m}}_k = \frac{1}{n_k} \sum_{\bx_i \in \Pi_k} \bx_i $
is the class mean of the $k$th class $\Pi_k$.
The between-class scatter matrix of the $i$th and $j$th classes is defined as
\begin{equation}
\bS_{ij} = (\overline{\mathbf{m}}_i-\overline{\mathbf{m}}_j)(\overline{\mathbf{m}}_i-\overline{\mathbf{m}}_j)^\T.
\end{equation}

Unlike LDA which seeks to minimize the average within-class pairwise distance, the within-class scatter measure used in WLDA is defined as
	\begin{equation}
	\rho_w = \max_{1 \leq k \leq c} \left\lbrace {\trace (\bW^\T \bS_k \bW)} \right\rbrace,
	\label{EQ:within}
	\end{equation}
which is the maximum of the within-class pairwise distances over all classes.

On the other hand, the between-class scatter measure used in WLDA is defined as
	\begin{equation}
	\rho_b = \min_{ 1 \leq i < j \leq c } \left\lbrace  {\trace (\bW^\T \bS_{ij} \bW)} \right\rbrace.
	\label{EQ:between}
	\end{equation}
$\rho_b$ uses the minimum of the pairwise distances between class means,
instead of the average of distances between each class mean and the sample mean employed in LDA.

The optimality criterion of WLDA is defined as to maximize
the ratio of the between-class scatter measure to the within-class scatter measure:
	\begin{subequations}
	\label{EQ:WLDA}
	\begin{align}
	\max_{\bW}  &\quad \frac{\rho_b}{\rho_w},  \\
     \sst &\quad   \bW^\T \bW = \eyes_r.
	\end{align}
	\end{subequations}

As stated in~\cite{Zhang2010NIPS}, this problem \eqref{EQ:WLDA} is not easy to optimize with respect to $\bW$, so a new variable $\bZ=\bW \bW^\T \in \Real^{d \times d}$ is introduced, and the problem \eqref{EQ:WLDA} is formulated as a metric learning problem.

	\begin{theorem}
	\label{THEO1}
	Define sets $\Omega_1 = \{ \bW\bW^\T\!: \bW^\T\bW=\eyes_r, \bW\in\Real^{d\times r}\} $,
	and $\Omega_2 = \{\bZ\!: \bZ = \bZ^\T,  \trace(\bZ) = r,
     \mathbf{0} \preccurlyeq \bZ \preccurlyeq \eyes_d \}$.
	Then $\Omega_1$ is the set of extreme points of $\Omega_2$, and $\Omega_2$ is the convex hull of $\Omega_1$.
	\end{theorem}

This theorem has been widely used and its proof can be found in~\cite{Overton1992}.
According to Theorem~\ref{THEO1}, the orthonormal constraint on $\bW$ can be relaxed to convex constraints on $\bZ=\bW \bW^\T$,
and the problem \eqref{EQ:WLDA} can be relaxed to the following problem defined on a \psd variable $\bZ$~\cite{Zhang2010NIPS}:
	\begin{subequations}
    \label{EQ:REL}
	\begin{align}
	\max_{\bZ} &\quad \frac{\min_{1 \leq i < j \leq c} \{ \trace(\bS_{ij} \bZ) \}}{\max_{1 \leq k \leq c}
		   \{ \trace(\bS_{k} \bZ) \}},  \\
	\sst &\quad  \trace(\bZ) = r,  \\
	     &\quad  \mathbf{0} \preccurlyeq \bZ \preccurlyeq \eyes_d.
	\end{align}
	\end{subequations}

Once the optimal solution $\bZ^\star$ is obtained,
the optimal $\bW^\star$ for problem \eqref{EQ:WLDA} can be recovered using the top $r$ eigenvectors of $\bZ^\star$.

In~\cite{Zhang2010NIPS}, Zhang \etal proposed two methods to solve \eqref{EQ:WLDA}, as we stated in Section~\ref{sec:intro}. In the first one, an iterative algorithm was presented to solve the relaxed problem \eqref{EQ:REL}, where an SDP problem needs to be solved at each step by standard SDP solver. This method is not scalable to problems with high dimensionality or large training data points. The second one is based on a greedy approach, which cannot guarantee to find a globally optimal solution.

Hence, in the next section, we will describe our algorithm (so called \wlda) of finding the transformation matrix $\bZ=\bW \bW^\T$ that maximizes \eqref{EQ:REL}, and demonstrate how to solve it using an efficient approach.

\section{A Fast SDP Approach to WLDA}
\label{SEC:SDP}
In this section, problem \eqref{EQ:REL} is firstly reformulated into a sequence of SDP optimization problems based on bisection search.
Then, a Frobenius norm regularization is introduced and the SDP problem in each step is solved through Lagrangian dual formulation.
With this \wlda \xspace method, the global optimum can be acquired for the relaxed problem \eqref{EQ:REL}. The computational complexity can be reduced as well by solving the dual problem using quasi-Newton methods, compared with solving the primal problem directly using interior-point based algorithm.

\subsection{Problem Reformulation}

By introducing an auxiliary variable $\delta$, problem \eqref{EQ:REL} can be rewritten as
         \begin{subequations}
         \label{EQ:lab8}
         \begin{align}
           \max_{\delta, \bZ} &\quad \delta,    \label{EQ:lab8a}   \\
           \sst &\quad \trace (\bS_{ij} \bZ)
                \geq \delta \trace (\bS_{k} \bZ), \,\, \forall 1 \leq i < j \leq c, 1 \leq k \leq c,    \label{EQ:lab8b} \\
               &\quad  \trace (\bZ) = r,          \\
               &\quad \,\, \mathbf{0} \preccurlyeq \bZ \preccurlyeq \eyes_d.
         \end{align}
         \end{subequations}

There are two variables $\delta, \bZ$ to be optimized in problem \eqref{EQ:lab8}, but we are interested in finding $\bZ$ that can maximize $\delta$.
Problem \eqref{EQ:lab8} is clearly non-convex with respect to $\delta$ and $\bZ$ since the constraint \eqref{EQ:lab8b} is not convex. However, noting that \eqref{EQ:lab8b} will become linear if $\delta$ is given, we employ the bisection search strategy and convert the optimization problem
\eqref{EQ:lab8} into a set of convex feasibility problems, by which the global optimum can be computed effectively.

Let  $\delta^\star$  denote  the  optimal value of \eqref{EQ:lab8a}.
Given  $\delta^\circ  \in  \Real$,  if  the convex  feasibility  problem
         \begin{subequations}
         \label{EQ:lab9}
         \begin{align}
         \mathrm{find} &\quad \bZ,         \label{EQ:lab9a}   \\
         \sst &\quad \trace  ( \bS_{ij} \bZ )  \geq  \delta^\circ \trace(\bS_k \bZ), \,\, \forall 1 \leq i < j \leq c, 1 \leq k \leq c,   \label{EQ:lab9b}  \\
              &\quad  \trace (\bZ) =  r,     \label{EQ:lab9c}   \\
              &\quad  \,\, \mathbf{0} \preccurlyeq \bZ \preccurlyeq \eyes_d,
         \label{EQ:lab9d}
         \end{align}
         \end{subequations}
 is feasible, then $ \delta^\star  \geq  \delta^\circ  $.
 Otherwise, if the above problem is infeasible, then $ \delta^\star < \delta^\circ $.

Algorithm~\ref{ALG:Bisection} shows the bisection search based optimization process.
Once a feasible solution $\bZ^\star$ is obtained which maximize $\delta$,
$\bZ^\star$ will be the globally optimal solution to the relaxed problem \eqref{EQ:REL}.
The optimal $\bW^\star$ for problem \eqref{EQ:WLDA} can be acquired using the top $r$ eigenvectors of $\bZ^\star$.

 \begin{algorithm}[t]
 \caption{Solving problem \eqref{EQ:lab8} by bisection search.}
 \label{ALG:Bisection}
 \begin{algorithmic}
    \REQUIRE
     $\delta_l$: the lower bound of $\delta$;  $\delta_u$: the upper bound of $\delta$; and the tolerance $\sigma > 0$.
    \WHILE{ $  \frac {\abs{\delta_u - \delta_l}} {\delta_l} > \sigma  $ }

    \STATE 1). $\delta^\circ  =  \frac{  \delta_l  +  \delta_u  }{2}  $.
    \STATE 2). Solve SDP feasibility problem~\eqref{EQ:lab9} %
    \IF{\eqref{EQ:lab9} is feasible}
    \STATE   Get the feasible solution $\bZ$, $\delta_l = \delta^\circ$;
    \ELSE
    \STATE   $\delta_u = \delta^\circ$.
    \ENDIF
    \ENDWHILE
    \STATE 3). $\delta^\star = \delta^\circ$, and save the corresponding $\bZ^\star$.
    \ENSURE  $\bZ^\star $,$\delta^\star$.
 \end{algorithmic}
 \end{algorithm}

\vspace*{-10pt}
\subsection{Lagrangian Dual Formulation}
Algorithm~\ref{ALG:Bisection} shows that an SDP feasibility problem needs to be solved at each step during the bisection search process.
Considering that standard interior-point SDP solvers have a computational complexity of $\mathcal{O}(m^3+md^3+m^2d^2)$, where $d$ is the dimension of input data, and $m$ is the number of constraints in SDP,
it becomes quite expensive for processing high-dimensional data. In this subsection, we reformulate the feasibility problem \eqref{EQ:lab9} into a Frobenius norm regularized SDP problem, which can be efficiently solved via its Lagrangian dual using first-order methods like quasi-Newton. The computational complexity will be reduced to $\mathcal{O}(d^{3})$.
The primal solution $\bZ^\star$ can then be calculated from the dual solution based on Karush\text{-}Kuhn\text{-}Tucker (KKT)~\cite[p.~243]{boyd2004convex} conditions.

The problem \eqref{EQ:lab9} can be expressed equivalently in the following form:
		\begin{subequations}
		\label{EQ:feasibilityX}
         \begin{align}
         \mathrm{find} & \quad \bX = \left[
						\begin{array} {cc}
						\bZ  & \mathbf{0} \\
						\mathbf{0}& \bQ
						\end{array}
						\right],            \\
         	\sst & \quad \trace  ( \bar{\bS}_{ijk} \bX ) \geq 0, \,\, \forall 1 \leq i < j \leq c, 1 \leq k \leq c,   \label{EQ:feasibilityX_cons1} \\
         	& \quad  \trace (\bar{\eyes}_d \bX) = r,   \label{EQ:feasibilityX_cons2} \\
         	& \quad  \trace (\bH_{st}^\T \bX ) =\eyes_d(s,t), \,\, \forall 1 \leq t \leq  s \leq d,  \label{EQ:feasibilityX_cons3}\\
         	& \quad \hspace{1mm}  \bX \succcurlyeq \mathbf{0},
         	\end{align}
         \end{subequations}
where
$\bar{\bS}_{ijk}=\left[
	\begin{array} {cc}
	\bS_{ij}-\delta^\circ \bS_k  & \mathbf{0} \\
	\mathbf{0}& \mathbf{0}
	\end{array}
	\right]$,
	$\bar{\eyes}_d = \left[
	\begin{array} {cc}
	\eyes_d  & \mathbf{0} \\
	\mathbf{0}& \mathbf{0}
	\end{array}
	\right] $, and
		$\bH_{st}(p,q)=
		\begin{cases}
		1 & p=s,q=t; \,\,\, \text{or} \; p=t,q=s; \,\,  \\ & \text{or} \; p=s+d,
		  q=t+d; \,\, \\ & \text{or} \; p=t+d,q=s+d;\\
		0 & \text{Otherwise}.
		\end{cases}$.

In the above formulation, the variable $\mathbf{0} \preccurlyeq \bZ \preccurlyeq \eyes_d$ is replaced by
$\bX = \left[
	\begin{array} {cc}
	\bZ  & \mathbf{0} \\
	\mathbf{0}& \bQ
	\end{array}
	\right] \succcurlyeq \mathbf{0} $,
where $\bQ = \eyes_d -\bZ $.
The constraints \eqref{EQ:feasibilityX_cons1} and \eqref{EQ:feasibilityX_cons2} correspond to \eqref{EQ:lab9b} and \eqref{EQ:lab9c}, respectively.
The constraints \eqref{EQ:feasibilityX_cons3} stem from the fact that $\bQ + \bZ = \eyes_d $.

\begin{proposition}
\label{P3.1}
Given $\delta^\circ \in \Real$, if the problem \eqref{EQ:lab9} and equivalently \eqref{EQ:feasibilityX} is feasible,
one feasible solution exists for the following semidefinite least-squares problem:
\begin{subequations}
\label{EQ:SDP}
	\begin{align}
	\min_\bX & \quad \frac{1}{2} \| \bX \|_F^2, \\
	\sst & \quad \trace  ( \bar{\bS}_{ijk} \bX ) \geq 0, \,\,\,\, \forall 1 \leq i < j \leq c, 1 \leq k \leq c,
		                          \label{EQ:SDP1}\\
	& \quad  \trace (\bar{\eyes}_d \bX) = r,   \label{EQ:SDP2}\\
	& \quad  \trace (\bH_{st}^\T \bX ) =\eyes_d(s,t), \,\, \forall 1 \leq t \leq  s \leq d,
		                               \label{EQ:SDP3} \\
	& \quad \hspace{1mm}  \bX \succcurlyeq \mathbf{0}, \label{EQ:SDP4}
	\end{align}
\end{subequations}

\end{proposition}

If the problem \eqref{EQ:SDP} is feasible, its optimal solution $\bX^\star$ can be used as a feasible solution to \eqref{EQ:feasibilityX},
and one solution $\bZ^\star$ to problem \eqref{EQ:lab9} can be acquired as well.

The problem \eqref{EQ:SDP} is a standard semidifinite least-square problem and can be solved readily by off-the-shelf SDP solvers.
However, as we mentioned before, the computational complexity is really high if we solve the primal problem directly by standard interior-point SDP solvers. It will greatly hamper the use of WLDA in large-scale problems. Thanks to the Frobenius norm regularization in the objective function of \eqref{EQ:SDP}, we can use Lagrangian dual approach to solve the problem easily.

Introducing the Lagrangian multipliers $\mathbf{u}  \in \Real^{\frac{1}{2}(c^3-c^2)}$, $v \in \Real$, $\bP \in \Real^{\frac{1}{2}(d^2+d)}$ corresponding to the constraints \eqref{EQ:SDP1}-\eqref{EQ:SDP3},
and a symmetric matrix $\bY \in \Real^{d \times d}$ corresponding to the \psd constraint \eqref{EQ:SDP4}, the following result can be acquired.

\begin{proposition}
\label{prop_dual}
The Lagrangian dual problem of \eqref{EQ:SDP} can be simplified in the following form:
\begin{equation}
         \begin{aligned}
         \min_{\mathbf{u}, v, \bP} \,\, \frac{1}{2} \|(\bA)_+\|^2_F -v r - \sum_{s=1}^{d} \bP_{ss},   \,\,\,\,\,\,       %
         \sst \,\, \mathbf{u} \geq \mathbf{0},
                                                  \label{EQ:dualS}
         \end{aligned}
\end{equation}
where
\begin{equation}
\label{EQA}
\bA = \sum_{i,j,k} \mathbf{u}_{ijk} \bar{\bS}_{ijk} + v \bar{\eyes}_d + \sum_{s,t} \bP_{st} \bH_{st}^\T.
\end{equation}
Furthermore,
the optimal solution to problem \eqref{EQ:SDP} is $\bX^\star =  (\bA^\star)_+ $, where $\bA^\star = \sum_{i,j,k} \mathbf{u}^\star_{ijk} \bar{\bS}_{ijk} + v^\star \bar{\eyes}_d + \sum_{s,t} \bP^\star_{st} \bH_{st}^\T$, which is calculated based on the optimal dual variables $\mathbf{u}^\star$, $v^\star$, and $\bP^\star$.

\end{proposition}

From the definition of $\bA^\star$ and the operator $(\cdot)_+$, $\bX^\star$ is forced to be \psd and block-diagonal, so the optimal solution $\bZ^\star$ to problem~\eqref{EQ:lab9}  can be acquired easily. In addition, it is noticed that the objective function of \eqref{EQ:dualS} is differentiable (but not necessarily to be twice differentiable), it allows us to solve the dual problem efficiently using first-order methods, such as quasi-Newton methods.

The gradient of the objective function in problem \eqref{EQ:dualS} is
$g(\mathbf{u}, v, \bP)=\left[ \!
						\begin{array} {l}
						\langle \bA_+, \,\, \bS_{ijk} \rangle,  \, \forall 1 \leq i < j \leq c, 1 \leq k \leq c\\
						\langle \bA_+, \,\,  \eyes_d \rangle  - r  \\
					    \langle \bA_+, \,\, \bH_{st}^\T \rangle - \theta,\, \forall 1 \leq t  \leq s \leq d
						\end{array} \!
						\right]$,
where 	$ \theta =
		\begin{cases}
		1 & s=t; \\
		0 & s \neq t.
		\end{cases}$

\subsection{Feasibility Condition}

As stated in Proposition~\ref{P3.1},
if \eqref{EQ:lab9} is feasible, a solution can be found by solving the problem \eqref{EQ:SDP}.
During running quasi-Newton algorithms to solve the problem \eqref{EQ:SDP},
an infeasibility condition of the problem \eqref{EQ:lab9} needs to be checked iteratively, which is presented here:

	\begin{proposition}
	\label{Feability}
	If the following conditions are satisfied
     	\begin{equation}
      	\label{EQ:feasCon}
        \begin{cases}
      	 \frac {\| (\bA)_+ \|_F} {\abs{v r + \sum_{s=1}^{d} \bP_{ss}}} < \epsilon, \\
        v r + \sum_{s=1}^{d} \bP_{ss} > 0,
      	\end{cases}
      	\end{equation}
	then the problem
	\eqref{EQ:lab9} is considered as infeasible.
	\label{THEO3}
	\end{proposition}

This infeasibility condition can be deduced from a general conic feasibility problem presented in~\cite{Henrion2009}. Explanations are presented in detail in the appendix. %
We check this condition at each iteration of quasi-Newton algorithms. $\bA_+$ is evaluated during calculating the dual objective function, so it will not bring extra computational cost.
Once the condition \eqref{EQ:feasCon} is satisfied, problem \eqref{EQ:lab9} (equivalently \eqref{EQ:feasibilityX}) is not feasible and quasi-Newton algorithms will be stopped.
Otherwise, quasi-Newton algorithms keep running until convergence, and then
a feasible solution $\bX^\star = (\bA^\star)_+$ to the problem \eqref{EQ:feasibilityX} is found.

\subsection{Solving the Feasibility Problem}
In this subsection, we summarize the procedure of solving the problem \eqref{EQ:lab9} by our fast SDP optimization algorithm.
It has been domenstrated that we can find the feasible solution to \eqref{EQ:lab9} by solving the dual problem \eqref{EQ:dualS} with quasi-Newton methods.
In this work, we use
L-BFGS-B~\cite{Zhu1997,morales2011remark}, a limited-memory quasi-Newton algorithm package, which can handle the problem with box constraints.
Here we only need to provide the callback function to L-BFGS-B, which calculates the objective function of \eqref{EQ:dualS} and its gradient.
The procedure of finding the feasible solution $\bZ$ is described in Algorithm~\ref{ALG:Optimize}.

         \begin{algorithm}[t]
         \caption{Optimization procedure for solving problem \eqref{EQ:lab9}.}
         \label{ALG:Optimize}
         \begin{algorithmic}
            \REQUIRE $\delta^\circ$, $\mathrm{flag}=1$.
            Initialize dual variables $\mathbf{u}$, $v$, $\bP$.

            \STATE 1). Solve the dual \eqref{EQ:dualS} using L-BFGS-B.
            \REPEAT
            \STATE 1.1). Calculate the objective and gradient of the objective function in \eqref{EQ:dualS}.
            \STATE 1.2). Check the feasibility condition \eqref{EQ:feasCon}:
                 \IF{the condition \eqref{EQ:feasCon} is satisfied}
                    \STATE $\mathrm{flag}=0$ and {\bf break}.
                 \ENDIF
            \STATE 1.3). Update dual variables $\mathbf{u}$, $v$, $\bP$.
            \UNTIL {L-BFGS-B is converged.}

            \STATE 2). Compute $\bX^\star$ by eigen-decomposition.
            \STATE 3). Decompose $\bZ^\star$ from $\bX^\star$.
            \ENSURE  $\bZ^\star$, $\mathrm{flag}$ ($1$: feasible, $0$: infeasible).
         \end{algorithmic}
         \end{algorithm}

\subsection{Computational Complexity Analysis}

The computational complexity of L-BFGS-B is $\mathcal{O}(Km)$, where $K$ is a moderate number between 3 to 20, $m = \frac{1}{2}(c^3-c^2)+1+\frac{1}{2}(d^2+d)$ is the problem size to be solved by L-BFGS-B, which is equal to the number of constraints in the primal SDP problem \eqref{EQ:SDP}. At each iteration of L-BFGS-B, the eigen-decomposition of a $2d \times 2d$ matrix is carried out to compute $\bA_+$, which is used to evaluate all the dual objective values, gradients, as well as the feasibility conditions \eqref{EQ:feasCon}. The computational complexity is $\mathcal{O}(d^{3})$. Hence, the overall computational complexity of our algorithm \wlda \xspace is $\mathcal{O}(Km+d^{3})$. Since $Km \ll d^{3}$, eigen-decomposition dominates the most computational time of \wlda, which is $\mathcal{O}(d^{3})$.
On the other hand, solving an SDP problem using standard interior-point methods needs a computational complexity of $\mathcal{O}(m^3+md^3+m^2d^2)$. Since $m > d$ in our case, our algorithm is much faster than interior-point methods, and can be used to large-scale problems.

%% file: document4_exp.tex
\section{Experiments}
\label{SEC:Exp}
In this section, experiments are performed to verify the performance of \wlda. We conduct comparisons between \wlda \xspace
and other methods on both classification performance and computational complexity.
The classification performance is contrasted between \wlda \xspace and LDA, LMNN, OLDA.
We also compare the performance of our \wlda \xspace with both optimization methods proposed by Zhang~\etal in~\cite{Zhang2010NIPS} (Zhang~\etal (SDP) and Zhang~\etal (SOCP) receptively). This can be used to verify the correctness of our algorithm. The computational complexity is compared between \wlda \xspace, standard interior-point algorithms to solve our SDP formulation (SDPT3~\cite{Toh99sdpt3} and SeDuMi~\cite{Sturm98usingsedumi}), Zhang~\etal (SDP) which uses SDPT3 as well, and Zhang~\etal (SOCP).

All algorithms are tested on a $2.7$GHz Intel CPU with $20$G memory.
The \wlda \xspace algorithm is implemented in Matlab, where the Fortran code of L-BFGS-B is employed to solve the dual problem \eqref{EQ:dualS}.
The Matlab routine ``eig'' is used to compute eigen-decomposition.
The tolerance setting of L-BFGS-B is set to default.
The tolerance $\sigma$ in Algorithm~\ref{ALG:Bisection} is set to $1e\!-\!3$, and the parameter $\epsilon$ in the feasibility condition~\eqref{EQ:feasCon} is set to $1e\!-\!3$.

\subsection{Experiments on UCI Datasets}
Some UCI datasets~\cite{Frank+Asuncion:2010} are used here firstly.
We perform $30$ random splits for each dataset, with $70\%$ as training samples
and $30\%$ as test samples. The classification performance is evaluated
based on $5$ nearest neighbour ($5$-NN) classifier.
For fair comparison with LDA, the final dimension is set to $(c-1)$.

The experimental results are presented in Table~\ref{TAB:Data1}, where the baseline results are obtained by
applying $5$-NN classifier on the original feature space directly. For each dataset, the experiment runs $30$ times,
and the error rate is reported by the mean error as well as the standard deviation.
The smallest classification
error is shown in bold. The results illustrate that WLDA gives smaller classification error rates
compared to other algorithms in most datasets. The classification results by our fast SDP solving algorithm \wlda \xspace and Zhang~\etal~(SDP) are quite similar, with small difference coming from numerical error during computation. %
The error rates calculated by Zhang~\etal~(SOCP) are sometimes quite different from that by Zhang~\etal~(SDP) and \wlda \xspace, \eg, on ``Heart'' and ``Waveform'' datasets, which results from different relaxation methods and optimization procedures employed.

In terms of computational speed, \wlda \xspace approach is much faster than other methods. Zhang~\etal~(SDP), which is also solved by standard interior-point algorithm SDPT3, is faster than Ours (SDPT3), because of different SDP problem formulations.
The merit of \wlda \xspace on computation is even more dramatic for high dimensional problems. For example, we compare the computational speeds of \wlda \xspace and SDPT3
on the datasets of ``Iris'' and ``Waveform'', which have the same number of classes.
\wlda \xspace is about $5$ times faster than Zhang~\etal~(SDP), and $20$ times faster than Ours~(SDPT3) on ``Iris'' which has $105$ training samples with the input dimension as $4$,
whereas it becomes $12$ times quicker than Zhang~\etal~(SDP), and $300$ times quicker than Ours~(SDPT3) on ``Waveform'' which has $3500$ training samples with the input dimension as $40$.
The computational time increases more significantly for SeDuMi with respect to input dimension.
Zhang~\etal~(SOCP) has no computational advantage on solving problems with few training samples and low dimensionality. The computational superiority appears when dimensionality increases, referring to the results on ``Waveform'' dataset, which is quicker than Zhang~\etal~(SDP). Because of the column-wise iteration solving method of Zhang~\etal~(SOCP), the computational complexity of Zhang~\etal~(SOCP) relates closely with the final dimension. That is why Zhang~\etal~(SOCP) is quite fast on ``Sonar'' and ``Ionosphere'' datasets, which set the final dimension to $1$. However, it still slower than \wlda.

        \begin{table*}[t] \small
        \begin{center}
	\caption{Test errors and computational time of different methods on UCI datasets with $5$-NN classifier. The test error is the average over $30$ random splits,
                 with standard deviation shown in the bracket.
                 The computational time is also the average over $30$ runs. \wlda \xspace is efficient in computation,
                 and gives comparable classification performance compared to other methods.}
\vspace*{-8pt}
\label{TAB:Data1}
	 \setlength{\belowcaptionskip}{1mm}
   \scalebox{0.75}
   {
         \begin{tabular}{llllllll}
         \hline
$ $ & Heart &  Waveform  &  Iris  &  Balance  &  Sonar  &  Ionosphere \\
         \hline
$\#$ Train  & $206$ & $3500$ &  $105$  &  $438$  &  $146$  &  $246$  \\
$\#$ Test  &  $88$  & $1500$  & $45$  &  $187$  &  $62$  &  $105$  \\
$\#$ Classes      & $5$   &     $3$   & $3$    &  $3$     & $2$     & $2$  \\
Input Dim. &  $13$  &  $40$  &  $4$  &  $4$  &  $60$  &  $34$  \\
Final Dim. &  $4$  &  $2$  &  $2$  &  $2$  &  $1$ &  $1$  \\
\hline
\multicolumn{7}{l}{\textbf{Error Rates} (\%)}  \\
\hline
Euclidean  & $45.58$ ($3.66$) & $18.44$ ($0.94$) & $3.26$ ($1.82$) & $15.03$ ($1.94$) & \textbf{$\mathbf{25.00}$ ($\mathbf{4.76}$)} & $16.19$ ($2.26$) \\
LDA        & $36.10$ ($3.14$) & $15.60$ ($0.65$) & $3.19$ ($1.62$) & $10.88$ ($1.85$) & $27.10$ ($3.94$) & \textbf{$\mathbf{15.43}$ ($\mathbf{2.34}$)} \\
LMNN       & $41.33$ ($3.43$) & \textbf{$\mathbf{14.27}$ ($\mathbf{0.92}$)} & $3.26$ ($1.62$) & $11.50$ ($6.25$) & $49.73$ ($6.17$) & $20.86$ ($3.40$) \\
OLDA       & $36.36$ ($3.52$) & $15.58$ ($0.63$) & $3.19$ ($1.72$) & $10.90$ ($2.20$) & $26.61$ ($3.25$) & $16.14$ ($3.23$) \\
Zhang~\etal~(SDP)  & \textbf{$\mathbf{35.38}$ ($\mathbf{4.09}$)}  & $15.49$ ($0.96$) & \textbf{$\mathbf{2.89}$ ($\mathbf{1.67}$)} & $10.86$ ($2.49$) & $27.10$ ($3.47$) & $15.70$ ($3.21$) \\
Zhang~\etal~(SOCP)  & $37.12$ ($3.68$) & $17.34$ ($2.10$) & $2.96$ ($1.78$) &  \textbf{$\mathbf{10.43}$ ($\mathbf{2.19}$)} & $26.83$ ($2.87$) & $15.70$ ($3.39$) \\
\wlda        & $35.57$ ($4.64$) & $15.47$ ($1.01$) & \textbf{$\mathbf{2.89}$ ($\mathbf{1.67}$)} & $10.80$ ($2.23$) & $26.94$ ($3.32$) & \textbf{$\mathbf{15.43}$ ($\mathbf{2.34}$)} \\
\hline
\multicolumn{7}{l}{\textbf{Computation Time} }  \\
\hline
Ours~(SDPT3)   & $57.3$s & $16$m$40$s & $8.7$s & $11.1$s & $53$m$2$s & $5$m$20$s \\
Ours~(SeDuMi)  & $23.5$s & $1$h$31$m & $3.7$s & $5.2$s & $12$h$30$m & $30$m$10$s \\
Zhang~\etal~(SDP)  & $16.5$s & $37.0$s & $2.3$s & $2.0$s & $21.6$s & $58.2$s \\
Zhang~\etal~(SOCP)  & $71.8$s & $36.0$s & $26.2$s & $27.9$s & $14.1$s & $13.9$s \\
\wlda        & \textbf{$\mathbf{9.9}$s} & \textbf{$\mathbf{3.0}$s} & \textbf{$\mathbf{0.4}$s} & \textbf{$\mathbf{0.9}$s} & \textbf{$\mathbf{3.9}$s} & \textbf{$\mathbf{1.2}$s} \\
\hline
         \end{tabular}
       }
         \end{center}
        \end{table*}

\begin{figure}[tb] \small
\centering
\subfigure[Sonar] { \label{fig:1a}
  \includegraphics[width=0.3\textwidth]{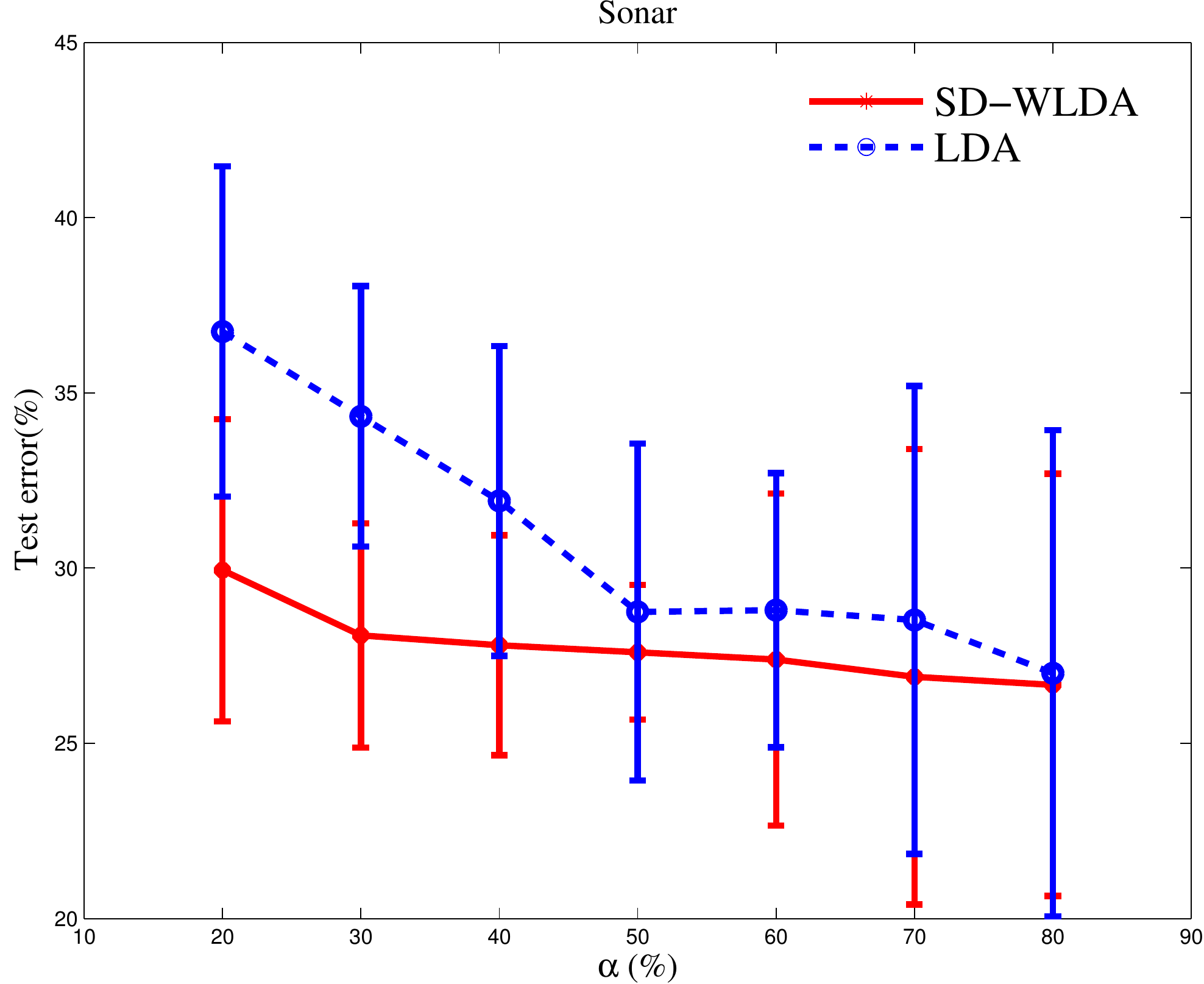}
}
\subfigure[Ionosphere] { \label{fig:1b}
\includegraphics[width=0.3\textwidth]{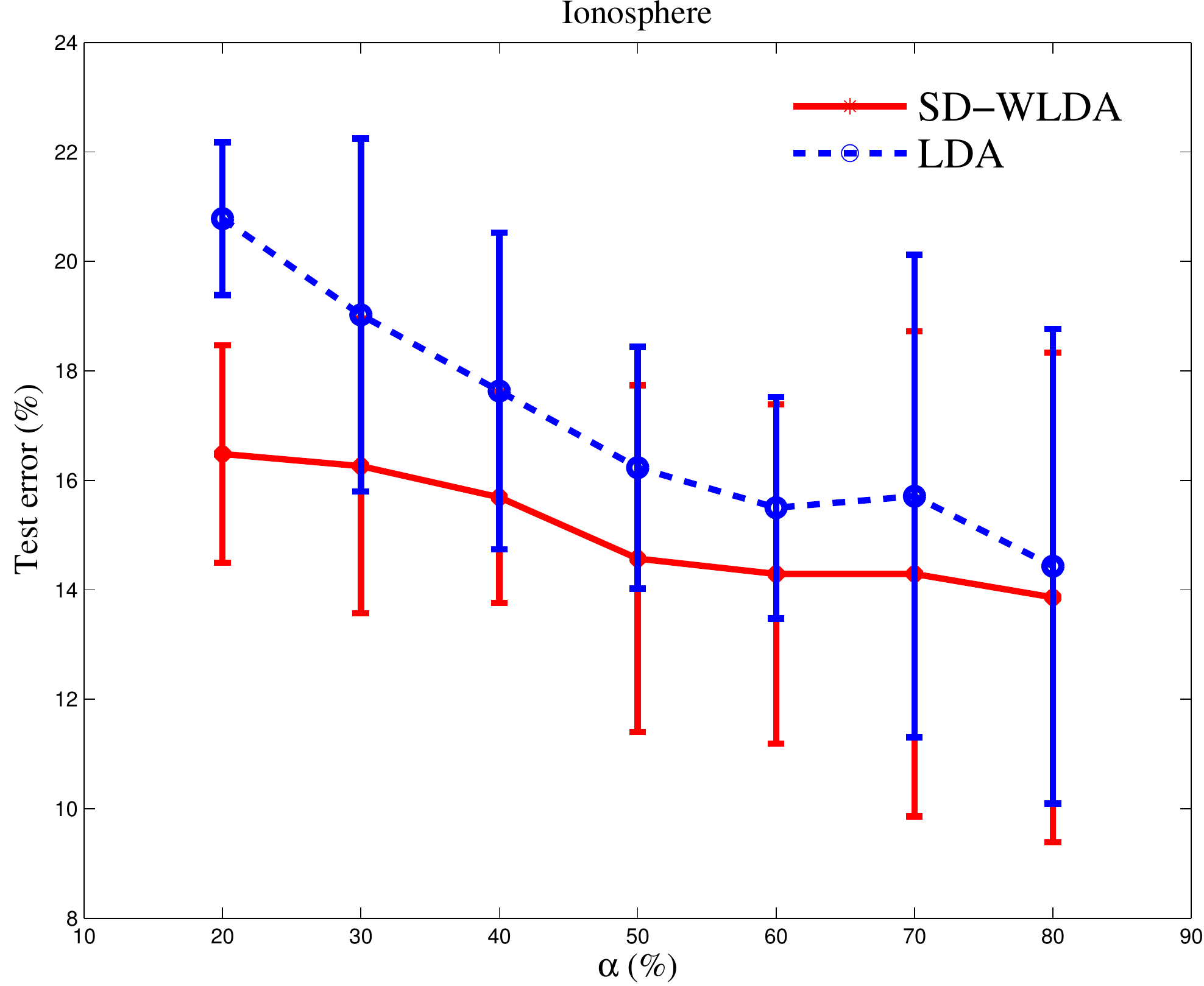}
}
\vspace*{-5pt}
\caption{Test errors of \wlda \xspace and LDA on ``Sonar'' and ``Ionosphere'' datasets, with different sizes of the training set. The test error is shown by marker,
which is the average error of $10$ trails for each split. The vertical bar represents the standard deviation.
\wlda \xspace produces significantly smaller errors than LDA with less number of training samples.}
\label{FIG:Card1}
\end{figure}

To prove the robust classification performance of \wlda, we change the ratio of training samples $\alpha$ from $20\%$ to $80\%$
on datasets ``Sonar'' and ``Ionosphere''. For each value of $\alpha$, we calculate the average test error as well as the standard deviation across $10$ trials by \wlda \xspace and LDA respectively.
The results in Fig.~\ref{FIG:Card1} demonstrate that \wlda \xspace is more superior than LDA when there is small number of training samples. %
This phenomenon illustrates that WLDA alleviates the dependence of
classification performance on large number of training samples.

LDA requires the data to map to at most $(c-1)$ dimension, while \wlda, which is based on an SDP optimization method, does not have such a restriction. Here we perform another experiment by \wlda \xspace on ``Heart'' dataset, with different final dimensions.
The result in Table~\ref{TAB:Data2} shows that $(c-1)$ is not the best
final dimension for ``Heart''. So \wlda \xspace algorithm is more flexible.

    \begin{table*}[tb] \small
	\begin{center}
	\caption{In contrast to LDA, \wlda \xspace can be used to project data to final dimension larger than $(c-1)$. The test error is by \wlda \xspace with $5$-NN classifier on ``Heart'' dataset using different final dimension. The test error is the average error of $10$ runs, with standard deviation in the bracket.}
	\vspace*{-8pt}
 \begin{tabular}{llllll}
	 \hline
	Final Dim.  & $4$ & $6$ & $7$ & $8$ & $9$  \\
	 \hline
	\wlda \xspace (\%) & $32.84$ ($3.95$) & $28.41$ ($2.94$) & $28.41$ ($2.29$) & $31.82$ ($3.88$) & $34.09$ ($4.84$) \\
	\hline
	 \end{tabular}
	 \label{TAB:Data2}
	 \end{center}
	\end{table*}

\subsection{Experiments on Face, Object and Letter Datasets}
As shown before, \wlda \xspace algorithm can be used to solve large-scale problems due to its efficiency on computation. In this section, experiments are carried out
on face, object and letter image datasets, which have high input dimension and more classes. The images are resized to different resolution before experiments, as shown in Table~\ref{TAB:Data3}. PCA has been applied beforehand to reduce the original dimension and also to
remove noises. The final dimension is still set to be $(c-1)$ for fair comparison to LDA.

\subsubsection{Face recognition}
four face databases are employed here.
ORL~\cite{ORL} consists of $400$ face images of $40$ individuals, each with $10$ images. We randomly choose $70\%$ of the samples for training and the remaining $30\%$ for test. The Yale dataset~\cite{Yale1997} contains 165 grey-scale images of $15$ individuals, $11$ images per subject. We split them into training and test sets by $7/3$ sampling as well. PIE dataset (Pose C29)~\cite{PIE} has $40$ subjects, and $24$ images for each individual. $80\%$ of the samples are chosen randomly for training. UMist dataset~\cite{UMist} contains 564 grayscale images of $20$ different people. We only use $30\%$ of the samples for training to test the performance of \wlda.

Experimental results in Table~\ref{TAB:Data3} show that \wlda \xspace gives better classification performance for all datasets. The classification error rates by \wlda \xspace and Zhang~\etal~(SDP) are identical with each other, as PCA used before has already removed the noises, which proves the correctness of our algorithm. However, \wlda \xspace is much faster than Zhang~\etal~(SDP) method. For example, \wlda \xspace runs almost $14$ times faster than Zhang~\etal~(SDP) on Yale dataset. The error rates calculated by Zhang~\etal~(SOCP) are rather larger, which result from the non-globally optimal solution the algorithm reached. The computational superiority of Zhang~\etal~(SOCP) does not show up as well.

\begin{figure}[t] \small
\begin{center}
\subfigure[Time Vs. Number of Class.] {\label{fig:2a}
  \includegraphics[width=.35\textwidth]{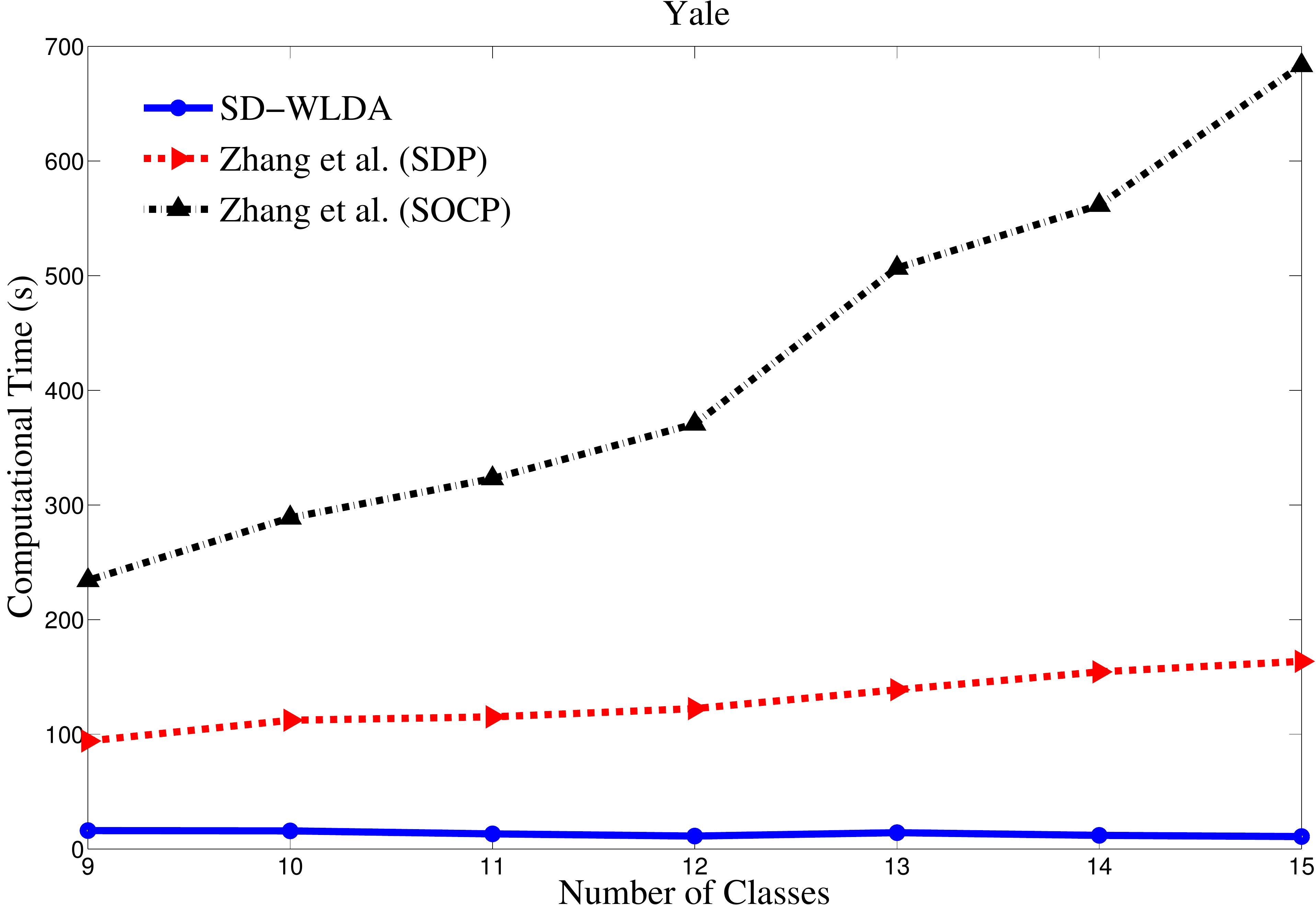}          %
}
\subfigure[Time Vs. Input Dim.] {\label{fig:2b}
  \includegraphics[width=.35\textwidth]{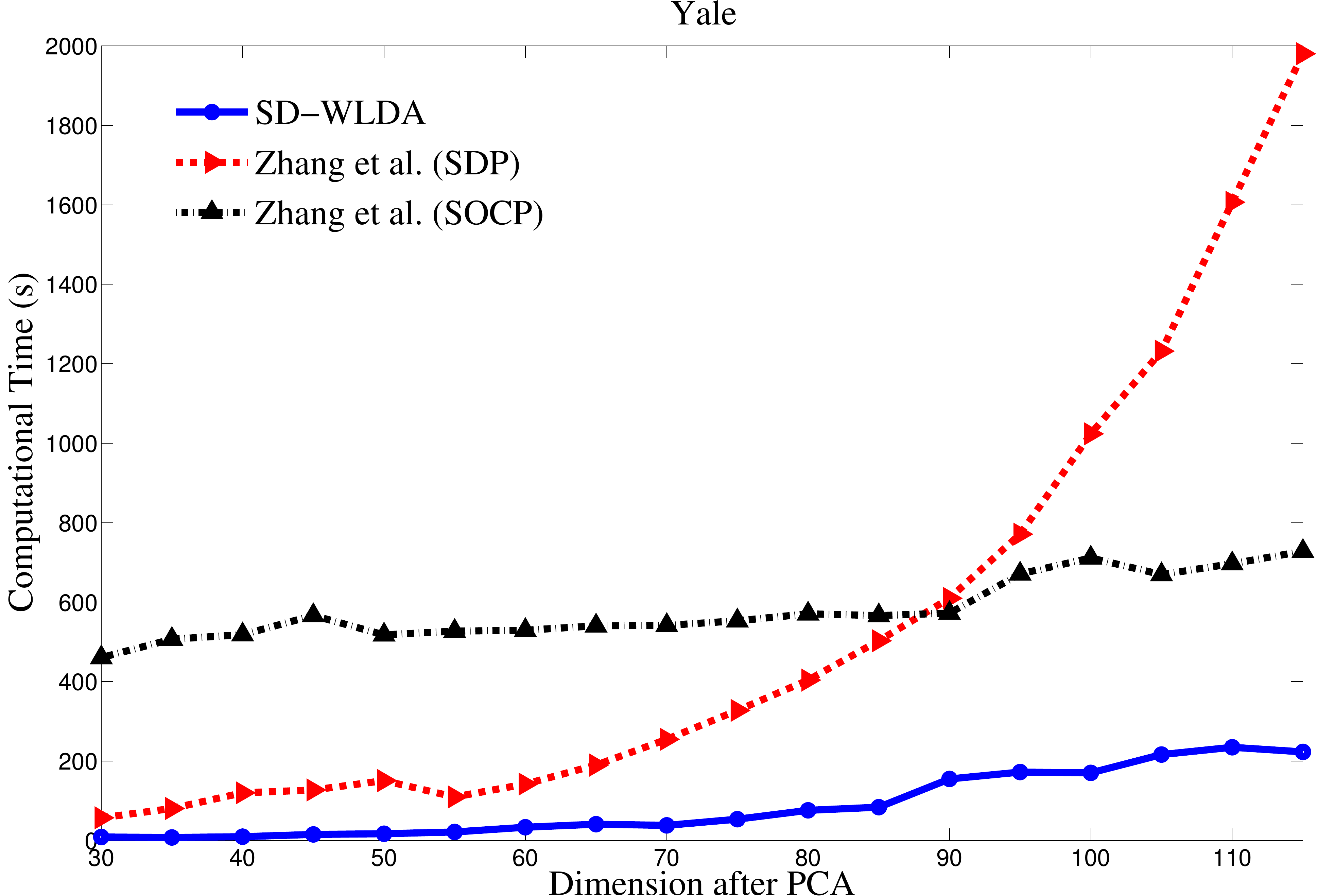}   %
}
\vspace*{-3pt}
\caption{Computational time of different SDP optimization methods with the increase of number of classes and input dimensionality, respectively. The computational time is the average time of $10$ trials for each setting. \wlda \xspace is more efficient than Zhang~\etal~(SDP) and Zhang~\etal~(SOCP) on computation. The computational complexity of \wlda \xspace is less sensitive to the increase of number of classes and input dimensionality, compared to Zhang~\etal~(SOCP) and Zhang~\etal~(SDP) respectively.}
\label{FIG:Card2}
\end{center}
\end{figure}

In order to illustrate the computational speed of \wlda \xspace and both methods in~\cite{Zhang2010NIPS} with respect to the number of classes and the input data dimension (here it refers to the dimension after PCA) respectively, more experiments are performed on Yale dataset.
Firstly, we set the dimension after PCA to be $50$, and change the number of classes from $9$ to $15$.
Experimental results in Fig.~\ref{fig:2a} demonstrate that compared with the other methods, the speed of \wlda \xspace is less sensitive to the increase of the amount of classes. Since the final dimension $r$ is set to be $(c-1)$, the computational complexity of Zhang~\etal~(SOCP) jumps up obviously with the increase of classes.
Secondly, we use all classes, and let the dimension after PCA change from $30$ to $115$.
Experimental results in Fig.~\ref{fig:2b} show that
the computational cost of \wlda \xspace rises up pretty slower with the growth of input dimension, in contrast to Zhang~\etal~(SDP). Zhang~\etal~(SOCP) becomes faster than Zhang~\etal~(SDP) when input dimension is larger than 90. This phenomenon certifies that Zhang~\etal~(SOCP) is more suitable for processing high dimensional datasets than Zhang~\etal~(SDP), as~\cite{Zhang2010NIPS} presented. However, this method cannot guarantee to find a global optimal solution.
Finally, we test on Yale dataset with an even high input dimension as $800$. Experimental results in Table~\ref{TAB:Data4} demonstrate that \wlda \xspace is absolutely faster than Zhang~\etal~(SDP), both of which lead to similar classification error rates. Although Zhang~\etal~(SOCP) is comparable with \wlda \xspace on computational time, the error rate it obtained is relatively bigger.

\begin{figure}
  \centering
  \includegraphics[width=.35\textwidth]{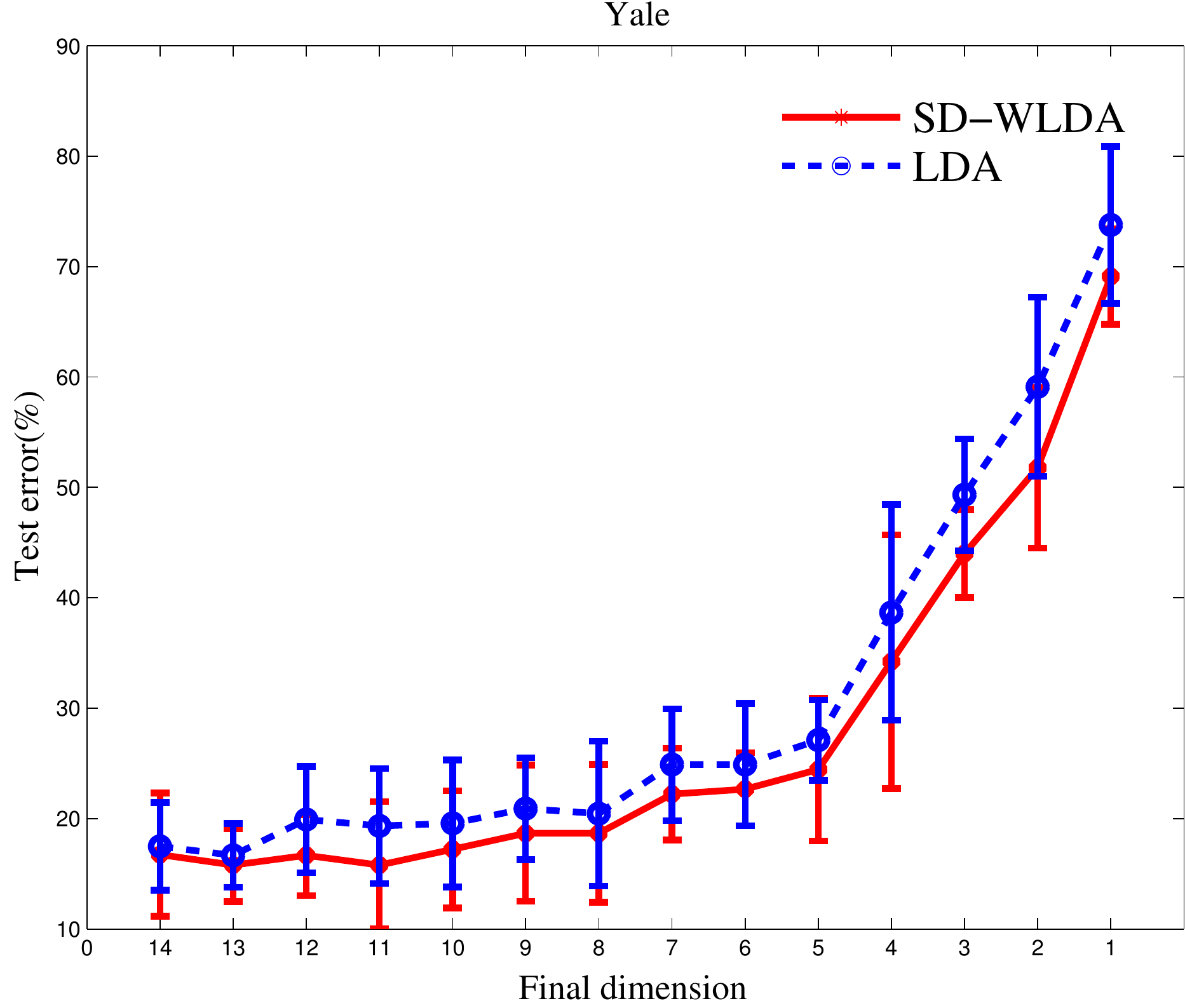}
\caption{Test errors of \wlda \xspace and LDA on Yale dataset with $5$-NN classifier using different final dimension ($120$ training samples, $45$ test samples, $15$ classes, and dimension after PCA is $50$). Test error is expressed by marker, which is the average error of $10$ trials. The vertical bar represents the standard deviation. \wlda \xspace always produces lower test errors than LDA in this experiment.}
\label{FIG:Card3}
\end{figure}

In addition, to show the superiority of \wlda \xspace on classification, another experiment is conducted using \wlda \xspace and LDA on Yale dataset,
with the dimension after PCA as $50$ and $15$ classes. We reduce the final dimension from $(c-1)$ to $1$.
The test results shown in Fig.~\ref{FIG:Card3} demonstrate that \wlda \xspace always gives lower test error than LDA, which further proves the good classification performance of \wlda.

 \begin{table*}[ht]  \small
    \begin{center}
    \caption{Test errors and computational time of different methods on lager-scale datasets with $5$-NN classifier. PCA is applied firstly. The test error is the average error with standard deviation in the bracket. The computational time is the average time too. \wlda \xspace gives better classification performance and faster computational speed than other algorithms.}
    \vspace*{-5pt}
    \label{TAB:Data3}
    \scalebox{.75}
{
     \begin{tabular}{l@{\hspace{0.1cm}}l@{\hspace{0.1cm}}l@{\hspace{0.1cm}}l@{\hspace{0.1cm}}l@{\hspace{0.1cm}}l@{\hspace{0.1cm}}l@{\hspace{0.1cm}}l@{\hspace{0.1cm}}l@{\hspace{0.1cm}}l}

         \hline
$ $ & ORL  &  Yale &  PIE & UMist  &  Coil20  &  Coil30 & ALOI & BA1 & BA2 \\
\hline
$\#$ Train & $280$ &  $120$  & $760$ & $174$ &  $580$  &  $390$  & $300$ & $120$ & $180$ \\

$\#$ Test &  $120$   & $45$  & $200$ & $401$ &  $860$  &  $360$ & $300$ & $270$ & $405$ \\

$\#$ Classes  & $40$     &  $15$  & $40$  & $20$ &  $20$  & $30$  & $25$ & $10$ & $15$ \\

Input Dim.  & $32\times32$   &  $32\times32$ & $64\times64$ & $112\times92$ & $64\times64$ & $64\times64$ & $58\times77$ & $20\times16$ & $20\times16$ \\

Dim. after PCA &  $100$   &  $50$ & $60$ & $100$ & $100$  &  $80$ & $100$ & $80$ & $80$ \\

Final Dim. &  $39$   &  $14$  & $39$ & $19$ &  $19$  &  $29$ & $24$ & $9$ & $14$ \\

$\#$ Runs   & $5$   & $10$   & $5$ & $10$ &  $10$  & $5$ & $5$ & $10$ & $10$ \\
         \hline
\multicolumn{7}{l}{\textbf{Error Rates} (\%) }  \\
\hline
Euclidean & $12.50$ ($3.39$) & $37.11$ ($5.35$) & $19.50$ ($2.12$) & $19.40$ ($2.58$)  & $7.09$ ($0.98$) & $9.89$ ($0.84$) & $4.33$ ($2.01$) & $18.07$ ($1.86$) & $23.46$ ($0.55$)\\

LDA       & $3.00$ ($2.01$) & $19.78$ ($4.74$) & $7.00$ ($2.12$) & $3.44$ ($0.76$) & $3.26$ ($1.29$) & $5.67$ ($1.53$) & \textbf{$\mathbf{1.67}$ ($\mathbf{0.84}$)} & $18.07$ ($2.30$) & $22.62$ ($2.27$)\\

LMNN      & $3.67$ ($1.26$) & $30.89$ ($11.01$) & $10.00$ ($0.71$) & $6.18$ ($2.48$) & \textbf{$\mathbf{2.91}$ ($\mathbf{0.35}$)} & $6.00$ ($1.14$) & $3.00$ ($0.98$) & $17.41$ ($3.39$) & $21.83$ ($1.63$)\\

OLDA       & $2.67$ ($1.24$) & $20.22$ ($5.59$) & $6.50$ ($2.21$) & $2.79$ ($0.41$) & $3.26$ ($0.99$) & $5.17$ ($0.87$) & $2.00$ ($0.72$) & $18.21$ ($1.94$) & $22.02$ ($1.95$)\\

Zhang~\etal~(SDP)     & \textbf{$\mathbf{1.83}$ ($\mathbf{0.91}$)} & \textbf{$\mathbf{18.44}$ ($\mathbf{3.64}$)} & \textbf{$\mathbf{6.25}$ ($\mathbf{1.43}$)} & \textbf{$\mathbf{2.94}$ ($\mathbf{0.65}$)} & ${3.14}$ (${0.51}$) & $4.22$ ($1.60$) & \textbf{$\mathbf{1.67}$ ($\mathbf{0.96}$)}  & \textbf{$\mathbf{16.22}$ ($\mathbf{1.73}$)} & \textbf{$\mathbf{21.48}$ ($\mathbf{2.10}$)}\\

Zhang~\etal~(SOCP)     & $6.67$ ($1.26$) & $28.44$ ($4.27$) & $9.50$ ($2.36$) & $8.58$ ($2.40$) & $5.56$ ($2.04$) & $5.28$ ($1.69$) & $4.13$ ($1.57$)  & $27.48$ ($2.93$) & $24.74$ ($1.24$)\\

\wlda     & \textbf{$\mathbf{1.83}$ ($\mathbf{0.91}$)} & \textbf{$\mathbf{18.44}$ ($\mathbf{3.64}$)} & \textbf{$\mathbf{6.25}$ ($\mathbf{1.43}$)} & \textbf{$\mathbf{2.94}$ ($\mathbf{0.65}$)} & ${3.14}$ (${0.51}$) & \textbf{$\mathbf{4.17}$ ($\mathbf{1.61}$)} & \textbf{$\mathbf{1.67}$ ($\mathbf{0.96}$)}  & \textbf{$\mathbf{16.22}$ ($\mathbf{1.73}$)} & ${21.58}$ (${2.23}$)\\
\hline
\multicolumn{7}{l}{\textbf{Computational Time (Once)} }  \\
\hline
Ours~(SDPT3)         & $>48$h & $2$h$41$m & $30$h$41$m  & $29$h$50$m & $29$h$10$m & $30$h$17$m & $>48$h & $5$h & $5$h$50$m \\

Ours~(SeDuMi)        & $>48$h  & $12$h$58$m & $>48$h & $>48$h & $>48$h & $>48$h & $>48$h & $27$h$47$m & $30$h$33$m \\

Zhang~\etal~(SDP)      & $1$h$4$m  & $2$m$44$s & $20$m & $13$m$30$s & $22$m$40$s & $15$m$52$s & $28$m$20$s & $2$m$38$s & $4$m$33$s \\

Zhang~\etal~(SOCP)     & $2$h$24$m  & $12$m$29$s & $1$h$45$m & $26$m & $1$h$25$m & $1$h$11$m & $52$m$40$s & $6$m$11$s & $10$m$51$s \\

\wlda        & \textbf{$\mathbf{4}$m$\mathbf{34}$s} & \textbf{$\mathbf{11.3}$s} & \textbf{$\mathbf{1}$m$\mathbf{44}$s} & \textbf{$\mathbf{2}$m$\mathbf{52}$s} & \textbf{$\mathbf{1}$m$\mathbf{57}$s} & \textbf{$\mathbf{2}$m$\mathbf{33}$s} & \textbf{$\mathbf{6}$m$\mathbf{31}$s} & \textbf{$\mathbf{28.3}$s} & \textbf{$\mathbf{29.3}$s}\\
\hline
         \end{tabular}
  }
       \end{center}
       \end{table*}

\begin{table*}[bt] \small
     \centering
	 \caption{Classification performance and computational speed of different methods on Yale dataset, with high input dimension. ($120$ Training samples, $45$ Test samples, $15$ Classes, Dimension after PCA is $800$, Final Dimension is $(c-1)$). \wlda \xspace presents faster computational speed than Zhang~\etal~(SDP) and better classification performance than Zhang~\etal~(SOCP).}
	   \vspace*{-8pt}
         \begin{tabular}{lcc}
         \hline
  & Error Rates (\%) &  Computational Time (Once) \\
\hline
\wlda  & $\mathbf{16.00}$ ($\mathbf{5.96}$) & \textbf{$\mathbf{36}$m$\mathbf{40}$s} \\
Zhang~\etal~(SDP) & $16.89$ ($5.35$) & $4$h$15$m \\
Zhang~\etal~(SOCP) & $22.22$ ($5.56$) & $41$m$20$s \\
\hline
         \end{tabular}
         \label{TAB:Data4}
\end{table*}

         \begin{table*}[t] \small
	\centering
	\caption{Test errors of \wlda \xspace and LDA on ORL dataset with $5$-NN classifier using different number of classes ($200$ Training samples, $200$ Test samples, Dimension after PCA is $60$, Final Dimension is $(c-1)$). \wlda \xspace presents better classification performance especially for large number of classes.}
	  \vspace*{-8pt}
         \begin{tabular}{lccccccc}
         \hline
$\#$ Classes $c$  & $5$ & $10$ & $20$ & $25$ & $30$ & $35$ & $40$ \\
         \hline
\wlda  (\%) & $\mathbf{2.22}$ ($\mathbf{4.97}$) & $4.00$ ($2.39$) & $5.50$ ($1.20$) & $5.12$ ($1.84$) & $\mathbf{3.87}$ ($\mathbf{1.52}$) & $\mathbf{4.57}$ ($\mathbf{1.81}$) & $\mathbf{3.00}$ ($\mathbf{1.97}$) \\
LDA (\%) & $\mathbf{2.22}$ ($\mathbf{4.97}$) & $\mathbf{3.75}$ ($\mathbf{1.98}$) & $\mathbf{4.75}$ ($\mathbf{2.31}$) & $\mathbf{4.32}$ ($\mathbf{2.57}$) & $4.40$ ($0.60$) & $4.91$ ($2.00$) & $4.50$ ($2.43$) \\
\hline
         \end{tabular}
         \label{TAB:Data5}
    \end{table*}

    \begin{table*}[bt] \small
     \centering
	 \caption{Test errors of \wlda \xspace and LDA on Coil20 dataset with $5$-NN classifier, using different number of classes ($580$ Training samples, $860$ Test samples, Dimension after PCA is $60$, Final Dimension is $(c-1)$). \wlda \xspace presents better classification performance especially for large number of classes.}
	   \vspace*{-8pt}
         \begin{tabular}{lcccccc}
         \hline
$\#$ Classes $c$  & $5$ & $8$ & $11$ & $14$ & $17$ & $20$ \\
        \hline
\wlda (\%) & $\mathbf{1.58}$ ($\mathbf{1.17}$) & $\mathbf{3.14}$ ($\mathbf{0.78}$) & $\mathbf{3.47}$ ($\mathbf{1.09}$) & $\mathbf{1.83}$ ($\mathbf{0.73}$) & $\mathbf{2.03}$ ($\mathbf{0.88}$) & $\mathbf{3.30}$ ($\mathbf{0.30}$) \\
LDA (\%) & $\mathbf{1.58}$ ($\mathbf{1.56}$) & $\mathbf{3.14}$ ($\mathbf{1.06}$) & $3.94$ ($1.43$) & $2.53$ ($1.24$) & $2.27$ ($0.95$) & $3.39$ ($0.64$) \\
\hline
         \end{tabular}
         \label{TAB:Data6}
      \end{table*}

\subsubsection{Object recognition} Three datasets are used here: Coil20~\cite{coil20}, Coil30~\cite{coil100}, and ALOI~\cite{Geusebroek20005}. Coil20 dataset contains $1440$ grey-scale images with black background for $20$ objects,
with each containing $72$ different images. Coil30 dataset consists of $750$ RGB images for $100$ objects. We choose the first $30$ objects and convert them into greyscale images in our experiment.
ALOI dataset consists of
$1000$ objects taken at varied viewing angles, illumination angles, etc..
We use the first $25$ objects here, with $24$ images for every object.
Different training and test splitting ratios are adopted for different
datasets in order to test the performance under different situations, which can be found in Table~\ref{TAB:Data3}.
The experimental results demonstrate that \wlda \xspace is better in computational speed.
Although OLDA produces the smallest classification error on Coil20 ($2.91\%$), \wlda \xspace gives a comparable result ($3.14\%$).
\subsubsection{Letter recognition} The Binary Alphadigits dateset~\cite{BA} is employed here.
The dataset BA1 contains $10$ digits of $0$ to $9$, and BA2 contains $15$ capital letters $A$ through $O$.
Experimental results in Table~\ref{TAB:Data3} present that WLDA produces better classification performance on both databases. Zhang~\etal~(SDP) and \wlda \xspace give similar classification results, while Zhang~\etal~(SOCP) lead to much larger error rates.
In terms of computational speed, \wlda \xspace runs $5$ times faster than Zhang~\etal~(SDP) on BA1,
and more than $9$ times faster than Zhang~\etal~(SDP) on BA2, which has more training samples and number of classes. This experiment demonstrates again that \wlda \xspace is more efficient in processing large-scale datasets.
\subsubsection{Classification performance regarding to number of classes}
It has been validated that \wlda \xspace can give smaller classification errors than LDA using a small size of training set.
In this section, we evaluate the classification performance of \wlda \xspace and LDA with respect to
the number of classes $c$. Take the datasets ORL and Coil20 as examples. The experimental settings for each dataset are shown in the captions of Table~\ref{TAB:Data5} and~\ref{TAB:Data6} respectively. We choose different numbers of classes from each dataset, and compare the test errors of \wlda \xspace and LDA.
The results in Tables~\ref{TAB:Data5} and~\ref{TAB:Data6} demonstrate that when the number of classes is small, \wlda \xspace and LDA have comparable classification results, whereas when the number of classes increases, \wlda \xspace shows better classification performance. %

%% file: document4_conclusion.tex
\section{Conclusion}
 \label{SEC:Con} 
 
 In this work, an efficient SDP optimization algorithm has been proposed to solve the problem of
 worst-case linear discriminant analysis. WLDA takes into account the worst-case 
 pairwise distance between and within classes, which achieves better
 classification performance than conventional LDA. In order to reduce the computational complexity so that it can be applied to large-scale problems, a fast algorithm has been presented by introducing 
 the Frobenius norm regularization, and its Lagrangian dual can be simplified. Using our algorithm,
 the global optimum can be obtained in $\mathcal{O}(d^{3})$ time. The algorithm is simple 
 to implement and much faster than conventional SDP solvers. Experimental results 
 on some UCI databases as well as face and object recognition tasks show
 the effectiveness on classification performance and the efficiency on computation of \wlda.

%% file: Appendix.tex
\renewcommand{\theequation}{A\arabic{equation}}
\renewcommand{\thetable}{A.\arabic{table}}
\renewcommand{\thefigure}{A\arabic{figure}}
\renewcommand{\thesection}{A.\arabic{section}}
\renewcommand{\thetheorem}{A\arabic{theorem}}

 \section{Proof of the Proposition~\ref{prop_dual}}
 This section presents the derivation of Proposition~\ref{prop_dual}.
 \begin{proof}
 With the Lagrangian dual multipliers $\mathbf{u}$, $v$, $\bP$ and $\bY$, the Lagrangian function of the primal problem \eqref{EQ:SDP} can be written as
 \begin{eqnarray}
 &&\mathrm{L}(\bX,\mathbf{u}, v, \bP, \bY) \notag \\
 & = &  \frac{1}{2} \| \bX \|_F^2 - \sum_{i,j,k} \mathbf{u}_{ijk} \trace(\bar{\bS}_{ijk} \bX) \notag \\
                            &&- v \trace(\bar{\eyes}_d \bX) + v r
                            - \sum_{s,t} \bP_{st} \trace( \bH_{st}^\T \bX) \notag \\
                            &&+ \sum_{s,t} \bP_{st} \eyes_d(s,t)
                            - \langle \bX, \bY \rangle          \notag \\
 &=& \frac{1}{2} \| \bX \|_F^2  - \langle \bX, \bY \rangle  + v r
                             + \sum_{s=1}^{d} \bP_{ss} - \langle \bA, \bX \rangle ,         \label{EQ:Lagrand2}
 \end{eqnarray}
 with $\mathbf{u} \geq \mathbf{0}$, $\bY \succcurlyeq \mathbf{0}$ and $\bA$ defined as \eqref{EQA}.

 Based on KKT conditions for convex problems,
 we have $\partial \mathrm{L}(\bX,\bu^\star, v^\star, \bP^\star, \bY^\star) / \partial \bX = 0$ at the $\bX^\star$,
 where $\bX^\star$, $\bu^\star$, $v^\star$, $\bP^\star$ and $\bY^\star$ stand for primal and dual optimal variables, respectively.
 Then the relationship between primal and dual optimal variables is:
     \begin{equation}
 	\label{EQ:Lagrand4}
 	\bX^\star = \bY^\star + \bA^\star.
 	\end{equation}

 Substituting the expression for $\bX$ back into the Lagrangian \eqref{EQ:Lagrand2}, the dual problem is formulated as
          \begin{subequations}
          \label{EQ:dual}
          \begin{align}
          \max_{\mathbf{u}, v, \bP, \bY} &\quad -\frac{1}{2} \|\bY + \bA\|^2_F +v r +\sum_{s=1}^{d} \bP_{ss},     \\
          \sst &\quad \,\, \mathbf{u} \geq \mathbf{0}, \bY \succcurlyeq \mathbf{0}.
          \end{align}
          \end{subequations}

 Given fixed $\mathbf{u}$, $v$, $\bP$, problem \eqref{EQ:dual} can be simplified to
          \begin{align}
          \min_{\bY} \,\, \|\bY + \bA\|^2_F, \quad
          \sst \,\, \bY \succcurlyeq \mathbf{0},
  \label{EQ:dualY}
          \end{align}
 which is equivalent to projecting the matrix $(-\bA)$ to the positive semidefinite cone.
The closed-form solution to \eqref{EQ:dualY} is:
 	\begin{equation}
 	\label{EQ:Yopt}
 	\bY^\star = (-\bA^\star)_+.
 	\end{equation}

 Substituting this solution $\bY$ back into the dual problem \eqref{EQ:dual}, the simplified dual problem can be expressed as \eqref{EQ:dualS} presented.

 Once the optimal solutions $\mathbf{u}^\star$, $v^\star$ and $\bP^\star$ are obtained by solving \eqref{EQ:dualS},
 we can obtain the primal optimal variable $\bX^\star$ based on \eqref{EQ:Lagrand4} and \eqref{EQ:Yopt}:
 	\begin{equation}
 	\bX^\star = (-\bA^\star)_+ + \bA^\star = (\bA^\star)_+.  \label{EQ:Xopt}
 	\end{equation}
 \end{proof}

 \section{Explanation on the Feasibility Condition}
 In this part, we will briefly review a feasibility condition to a conic feasibility problem described in~\cite{Henrion2009},
 and then extend it to our feasibility problem \eqref{EQ:lab9}.

 Consider a conic feasibility problem of finding a point $x \in \Real^n$ such that
 	\begin{equation}
 	\begin{cases}
 	Ax \geq b,  \\
 	x \in \mathcal{K},  \label{EQ:ConP}
 	\end{cases}
 	\end{equation}
 where $A \in \Real^{m \times n}$ and $b \in \Real^m$ are given.
 $Ax=b$ defines an affine subspace $\mathcal{A}$,
 and $\mathcal{K} \in \Real^n$ is a convex cone.



%
%

Defining the polar cone of $\mathcal{K}$ as the set of points whose projection into $\mathcal{K}$ is 0, \ie,
   \begin{equation}
   \mathcal{K}^\circ := \{ z \in \Real^n : z^\T x \leq 0, x \in \mathcal{K}\},
   \end{equation}
Henrion and Malick~\cite{Henrion2009} proposed the following proposition.

\begin{proposition}
If there exists a point $y \in \Real^m$ such that the following conditions
 	\begin{equation}
 	\label{EQ:Conic}
    \begin{cases}
 	A^\T y  \in \mathcal{K}^\circ,  \\
 	b^\T y  > 0,
 	\end{cases}
 	\end{equation}
are satisfied, there would be no feasible solution to \eqref{EQ:ConP}.
\end{proposition}


 Based on this proposition, we can get a feasible condition to our problem \eqref{EQ:lab9}. In our problem, $\mathcal{K}$ is the set of positive semidefinite matrices, and the polar cone of $\mathcal{K}$ is the set of negative semidefinite matrices. Then the feasibility problem
   	\begin{equation}
   	\label{EQ:feasy}
      \begin{cases}
    \bA \preccurlyeq  \mathbf{0}, \\
    v r + \sum_{s=1}^{d} \bP_{ss} > 0,
   	\end{cases}
   	\end{equation}
gives the certificate of infeasibility of problem \eqref{EQ:feasibilityX} which is equivalent to \eqref{EQ:lab9}. That is to say, if there is a feasible solution to \eqref{EQ:feasy}, there is no feasible solution to \eqref{EQ:lab9}.
We formalize this result in the following remark.

 	\begin{proposition}
 	\label{ApFeas}
 	(i) If problem \eqref{EQ:feasibilityX} is feasible, then \eqref{EQ:feasy} is infeasible;  \\
 	(ii) If there exists $\mathbf{u}$, $v$ and $\bP$ such that $(\bA)_+ = 0$, $ v r + \sum_{s=1}^{d} \bP_{ss} > 0$, then there is no feasibility solution to \eqref{EQ:feasibilityX}.
 	\end{proposition}

 \begin{proof}
  (i) Suppose that there is a feasible solution $\mathbf{u}$, $v$ and $\bP$ to \eqref{EQ:feasy}, and take a feasible point $\bX  \succcurlyeq \mathbf{0}$ of the problem \eqref{EQ:feasibilityX}. $\bA \preccurlyeq \mathbf{0}$ implies that $\langle \bA, \bX \rangle \leq 0$, \ie,
  \begin{equation}
  \label{Apendix1}
   \sum_{i,j,k} \mathbf{u}_{ijk} \langle \bar{\bS}_{ijk}, \bX \rangle  + v \langle \bar{\eyes}_d, \bX \rangle + \sum_{s,t} \bP_{st} \langle \bH_{st}^\T, \bX \rangle \leq 0.
  \end{equation}
  Observing that $\bX$ satisfying the constraints in \eqref{EQ:feasibilityX}, the above inequality \eqref{Apendix1} is equivalent to
  \begin{equation}
  \label{Apendix2}
   \sum_{i,j,k} \mathbf{u}_{ijk} \langle \bar{\bS}_{ijk}, \bX \rangle  + v r +\sum_{s=1}^{d} \bP_{ss}  \leq 0.
  \end{equation}
  Since $\trace  ( \bar{\bS}_{ijk} \bX ) \geq 0$, and $ v r +\sum_{s=1}^{d} \bP_{ss} >0$, the above inequality \eqref{Apendix2} cannot hold at all, which means there is no feasible solution to \eqref{EQ:feasy}.

  (ii)  $(\bA)_+ = 0$ is equivalent to $\bA \preccurlyeq 0$. Combining with the condition $ v r + \sum_{s=1}^{d} \bP_{ss} > 0$, \eqref{EQ:feasy} is feasible. Therefore problem \eqref{EQ:feasibilityX} would be infeasible according to the contrapositive of (i).

 \end{proof}

$(\bA)_+ = 0$ is also equivalent to $\|(\bA)_+\|_F= 0$.
Due to numerical reasons, the latter $\epsilon$ is adopted in the feasibility condition~\eqref{EQ:feasCon}.

\section{Data Visualization}

\begin{figure*}[t!]
\vspace{-0.0cm}
\centering
 \subfigure[PCA, Yale]{
\centering \includegraphics[type=pdf,ext=.pdf,read=.pdf,width=0.24\textwidth,height=0.195\textwidth]{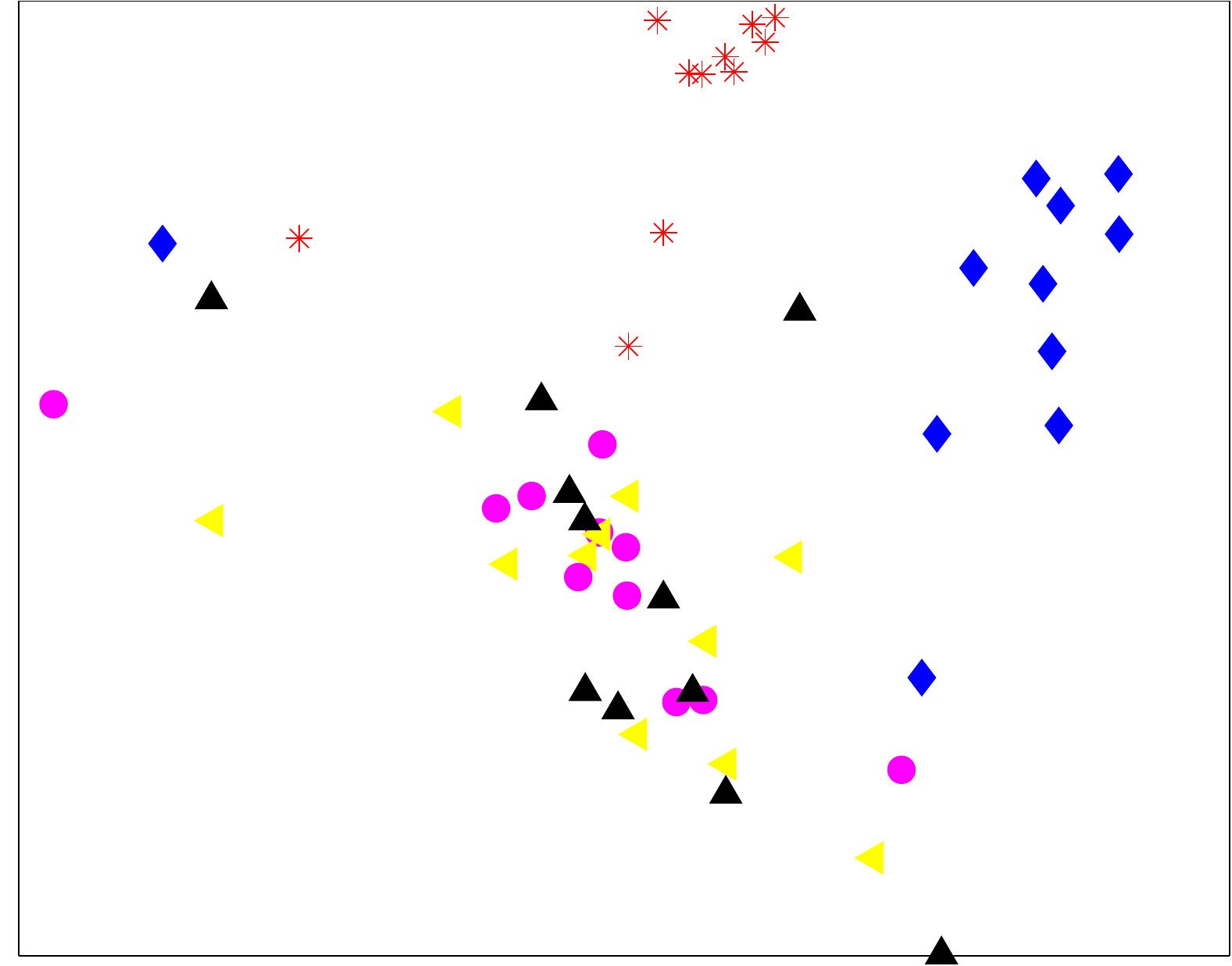}}\hspace{-0.25cm}
 \subfigure[PCA, ALOI]{
\centering \includegraphics[type=pdf,ext=.pdf,read=.pdf,width=0.24\textwidth,height=0.197\textwidth]{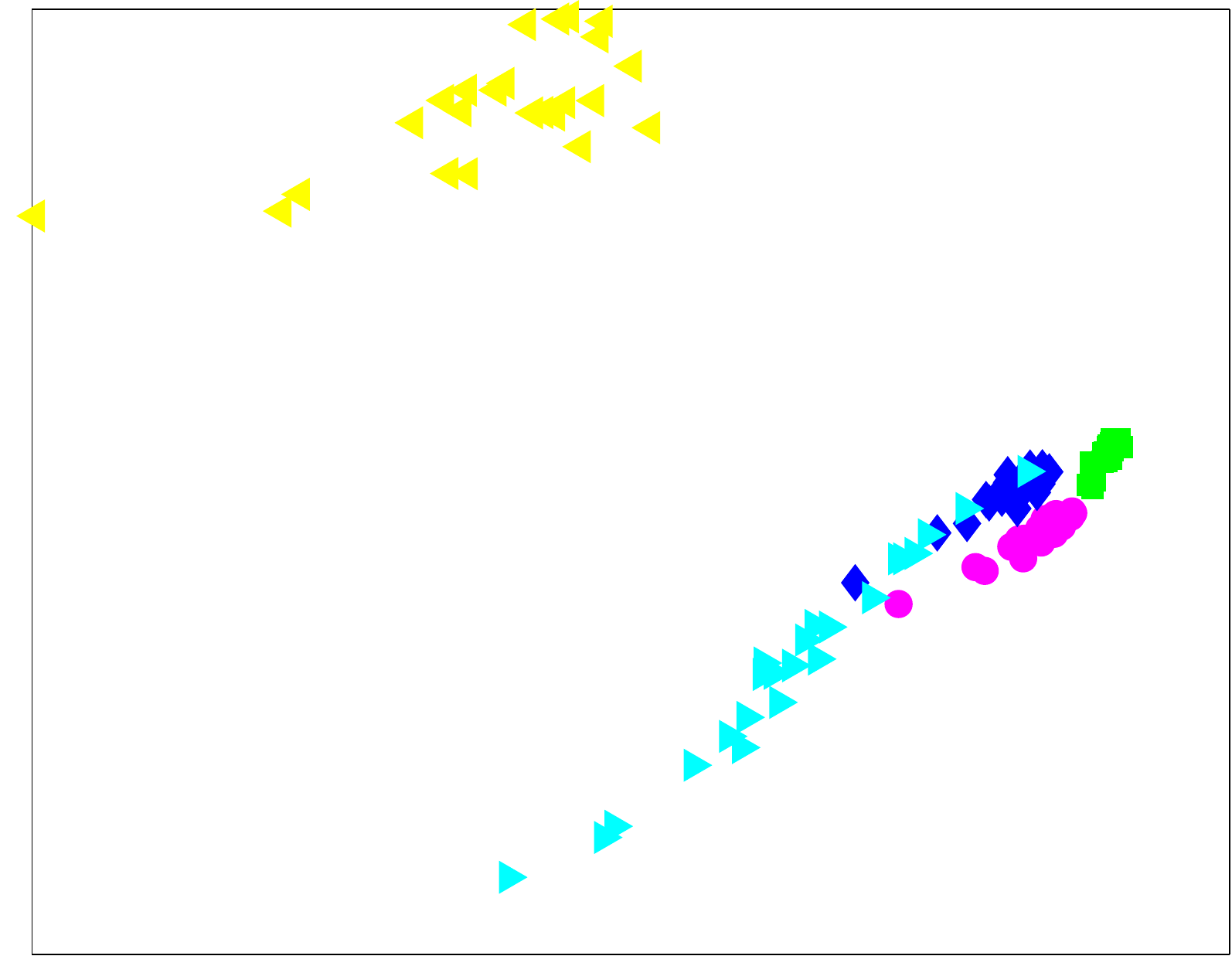}}\hspace{-0.75cm}
\hspace{0.4cm}
\subfigure[PCA, Coil20]{
\centering \includegraphics[type=pdf,ext=.pdf,read=.pdf,width=0.24\textwidth,height=0.195\textwidth]{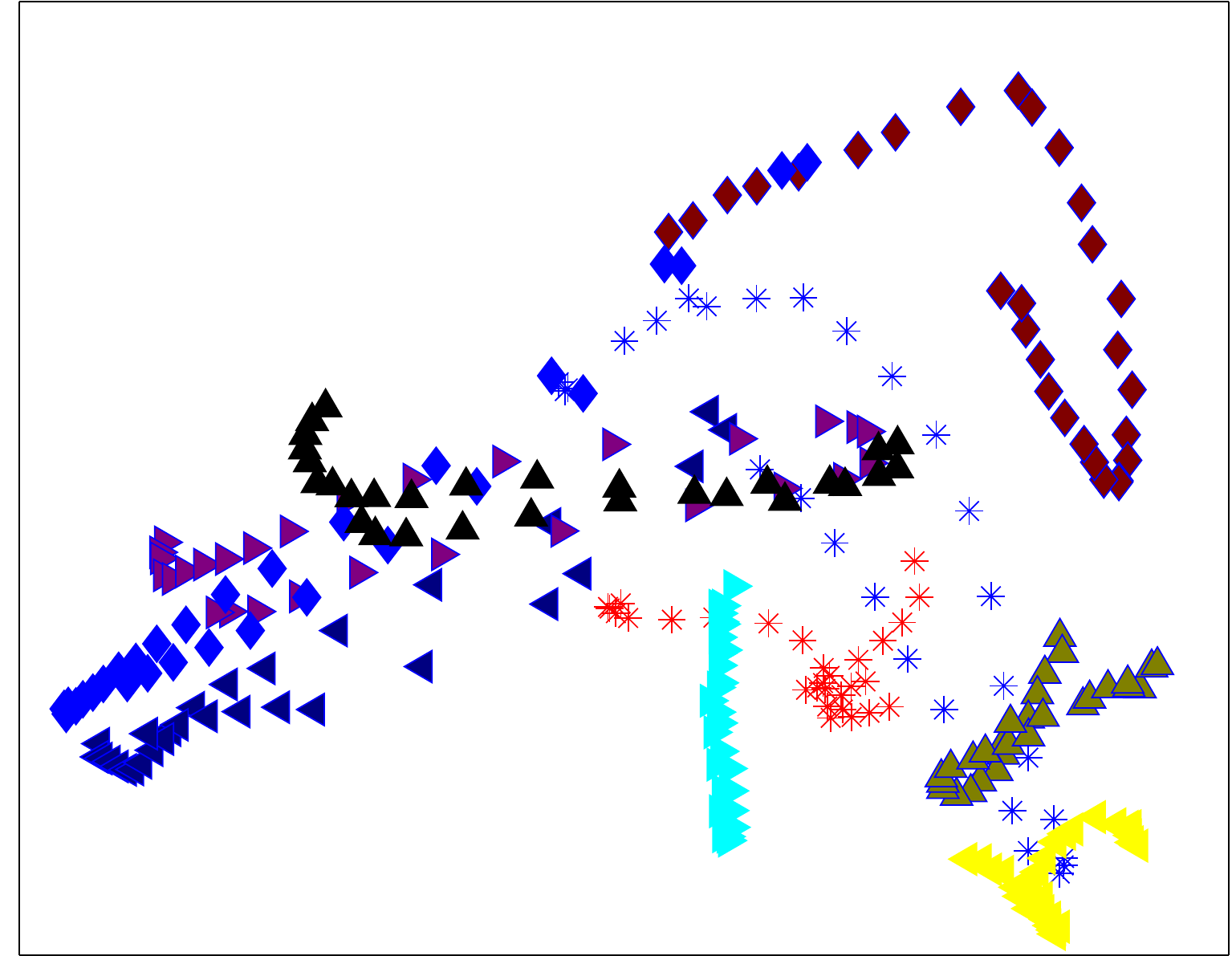}}\hspace{-0.2cm}
 \subfigure[PCA, BA1]{
\centering \includegraphics[type=pdf,ext=.pdf,read=.pdf,width=0.24\textwidth,height=0.195\textwidth]{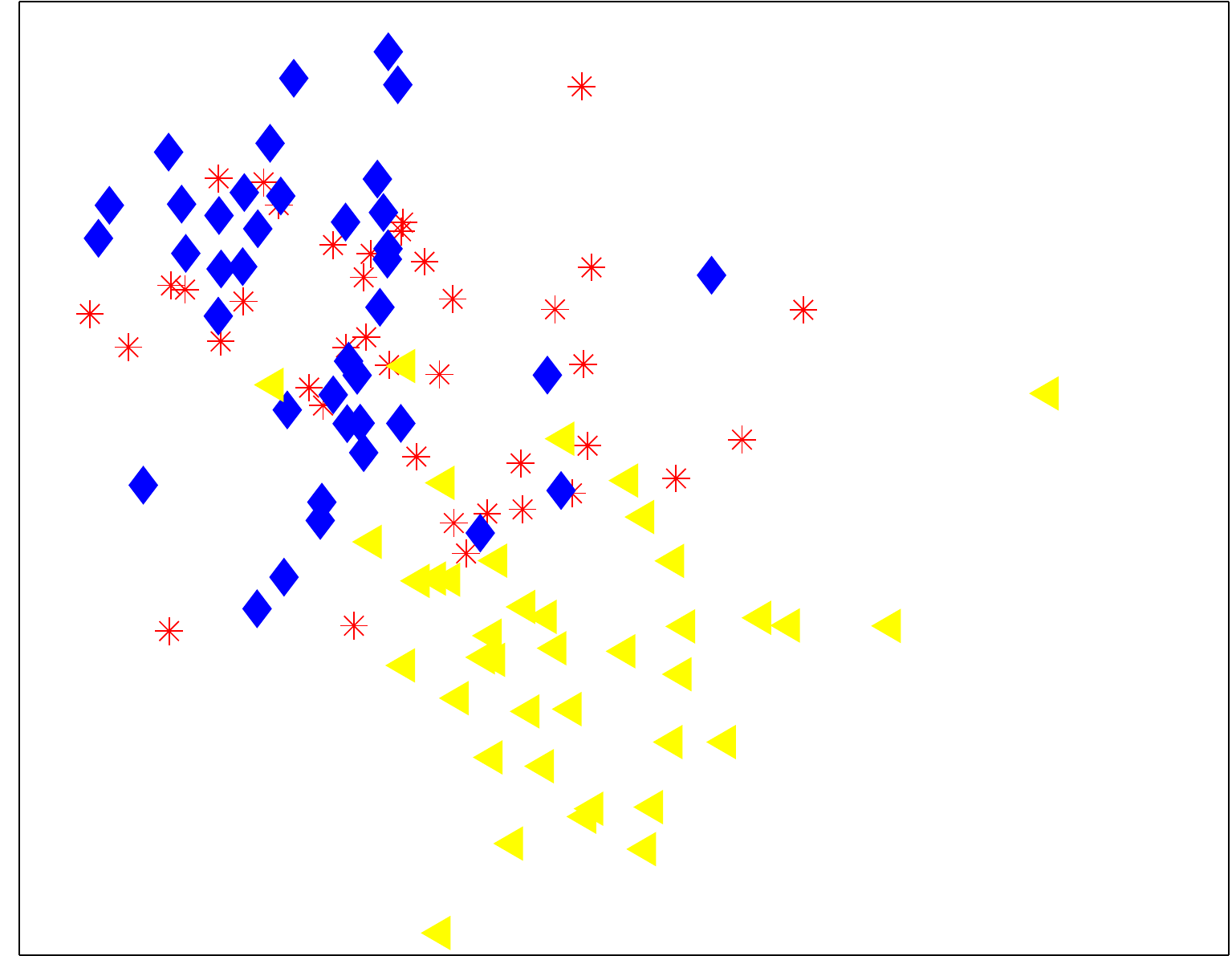}}
\\
 \subfigure[LDA, Yale]{
\centering \includegraphics[type=pdf,ext=.pdf,read=.pdf,width=0.24\textwidth,height=0.185\textwidth]{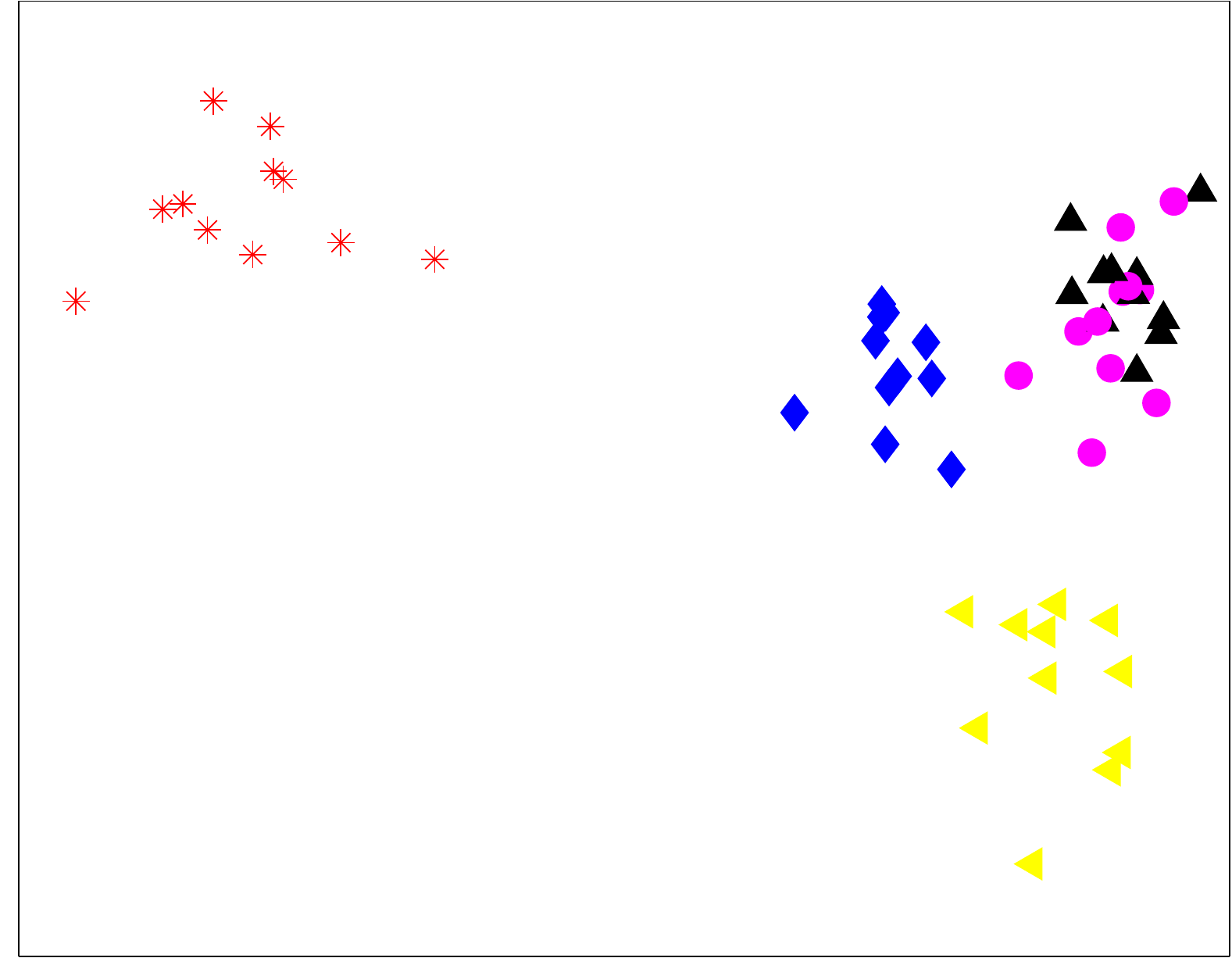}}\hspace{-0.2cm}
 \subfigure[LDA, ALOI]{
\centering \includegraphics[type=pdf,ext=.pdf,read=.pdf,width=0.24\textwidth,height=0.185\textwidth]{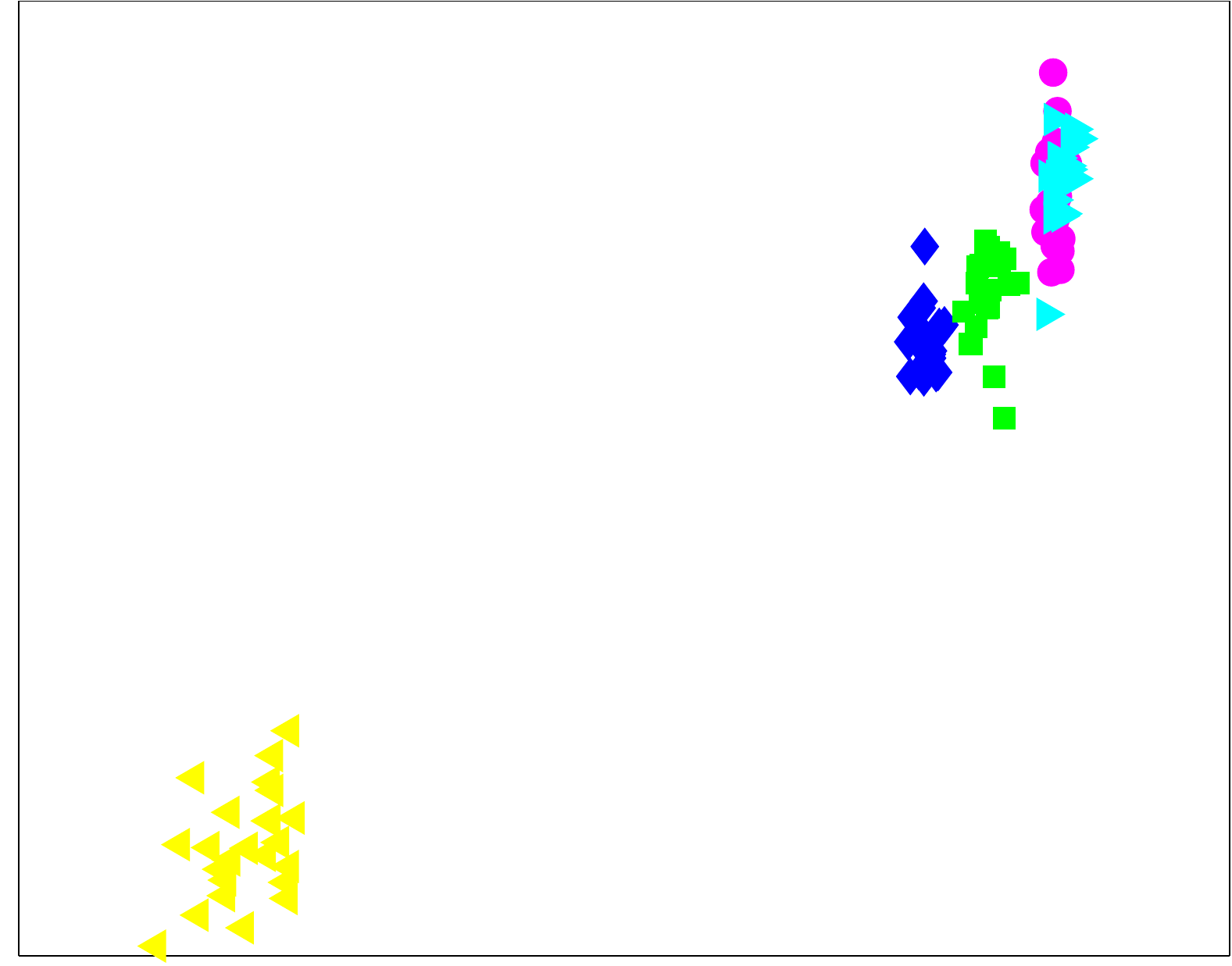}}\hspace{-0.2cm}
 \subfigure[LDA, Coil20]{
\centering \includegraphics[type=pdf,ext=.pdf,read=.pdf,width=0.24\textwidth,height=0.188\textwidth]{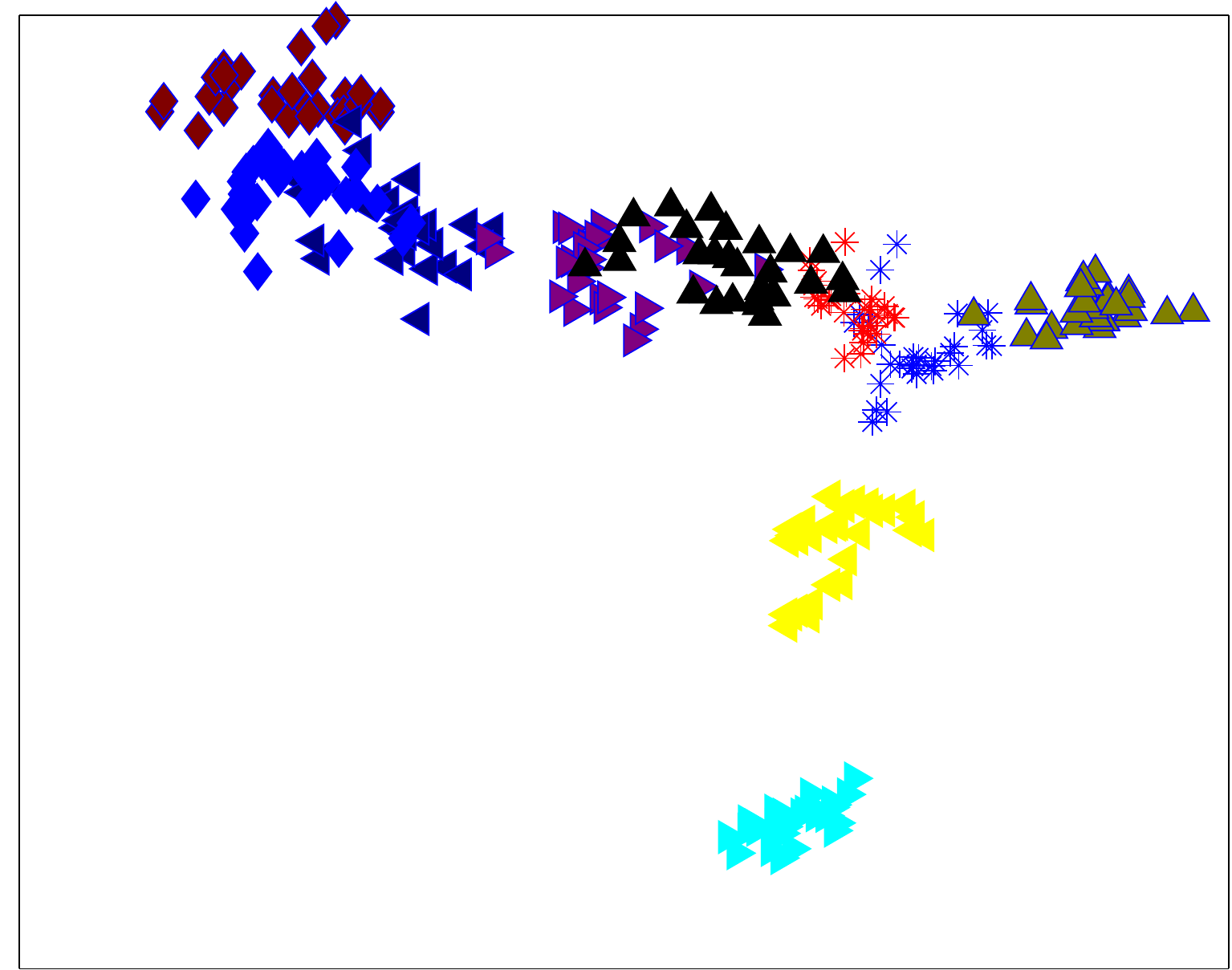}}\hspace{-0.2cm}
 \subfigure[LDA, BA1]{
\centering \includegraphics[type=pdf,ext=.pdf,read=.pdf,width=0.24\textwidth,height=0.185\textwidth]{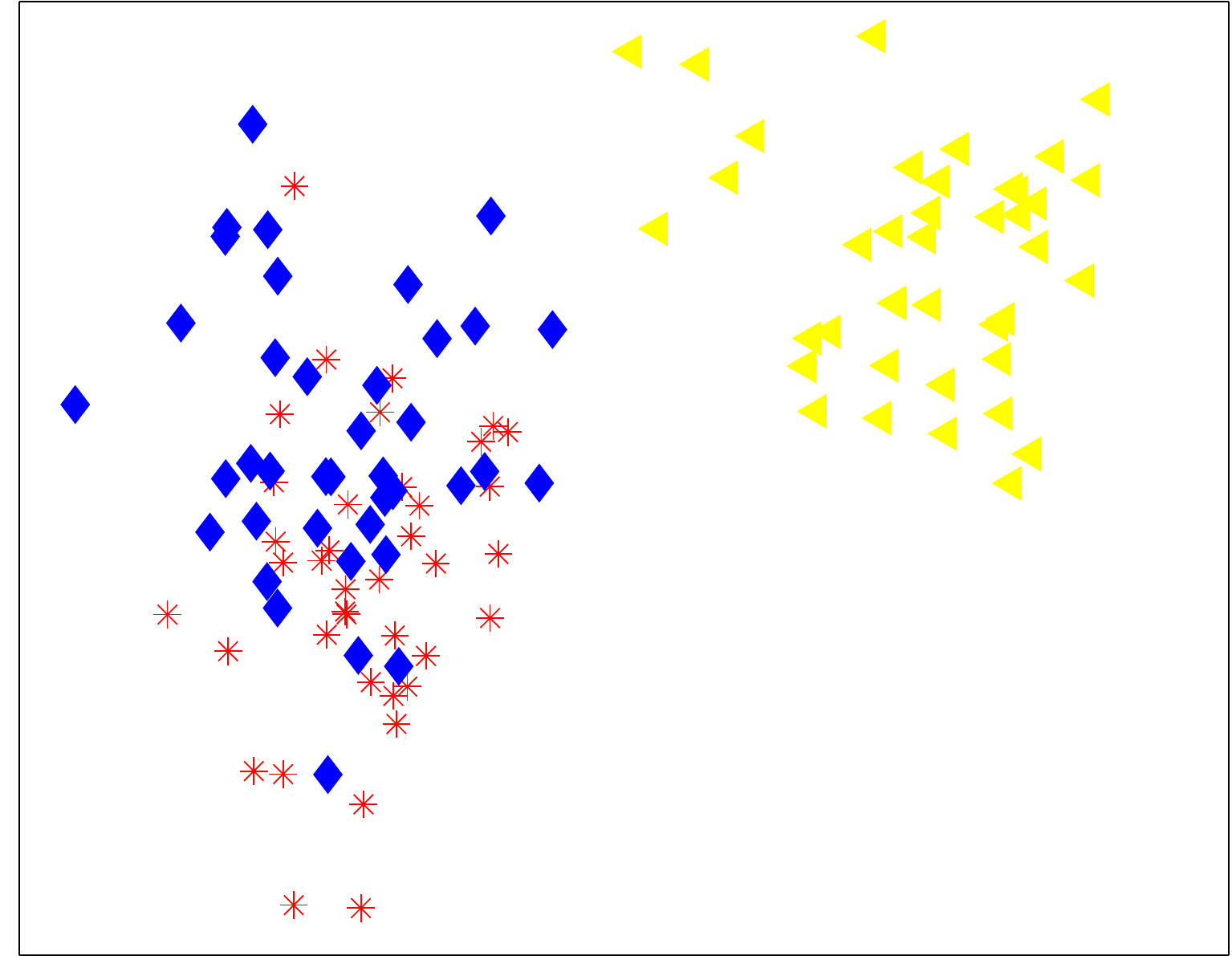}}
\\
 \subfigure[OLDA, Yale]{
\centering \includegraphics[type=pdf,ext=.pdf,read=.pdf,width=0.24\textwidth,height=0.187\textwidth]{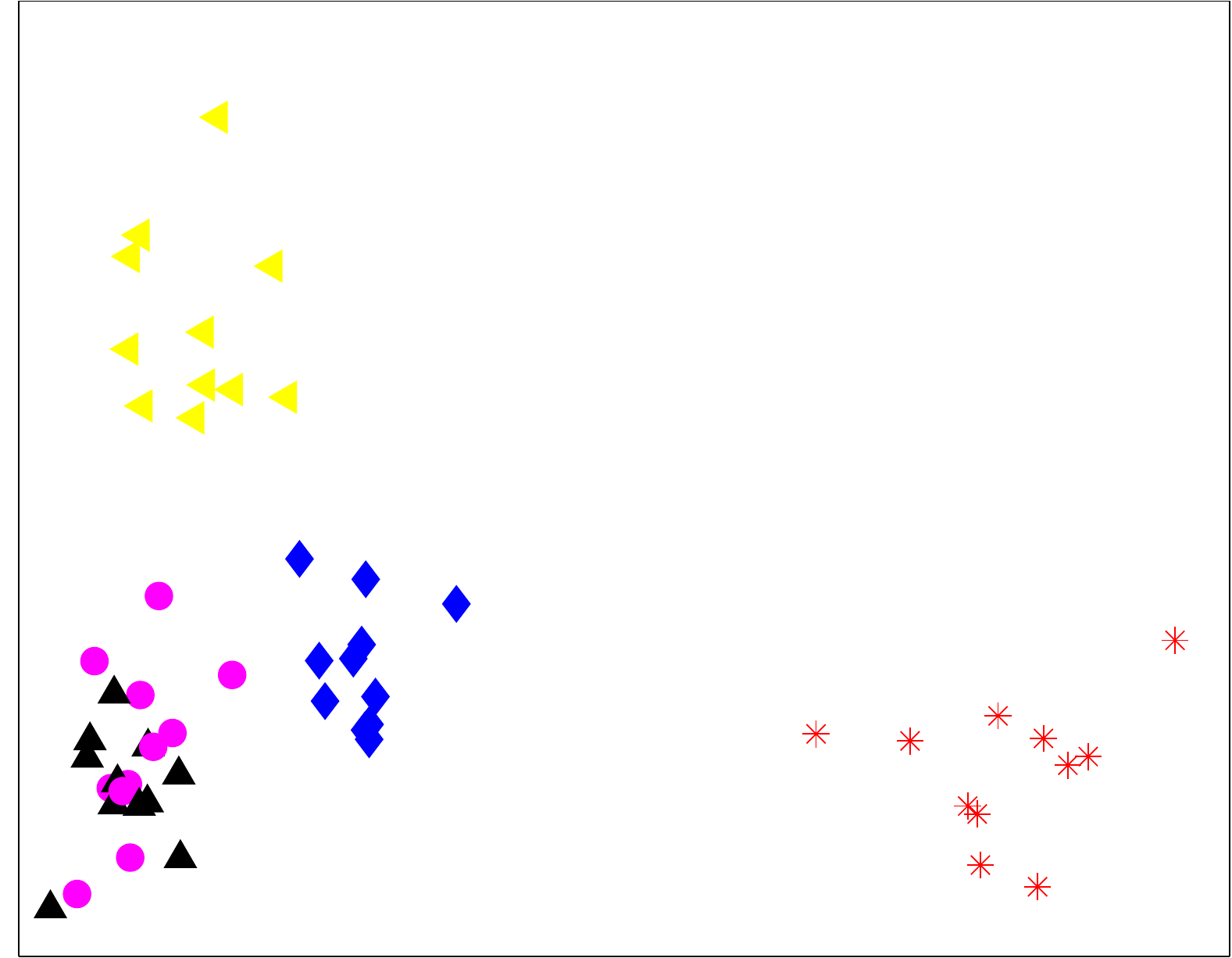}}\hspace{-0.2cm}
 \subfigure[OLDA, ALOI]{
\centering \includegraphics[type=pdf,ext=.pdf,read=.pdf,width=0.24\textwidth,height=0.187\textwidth]{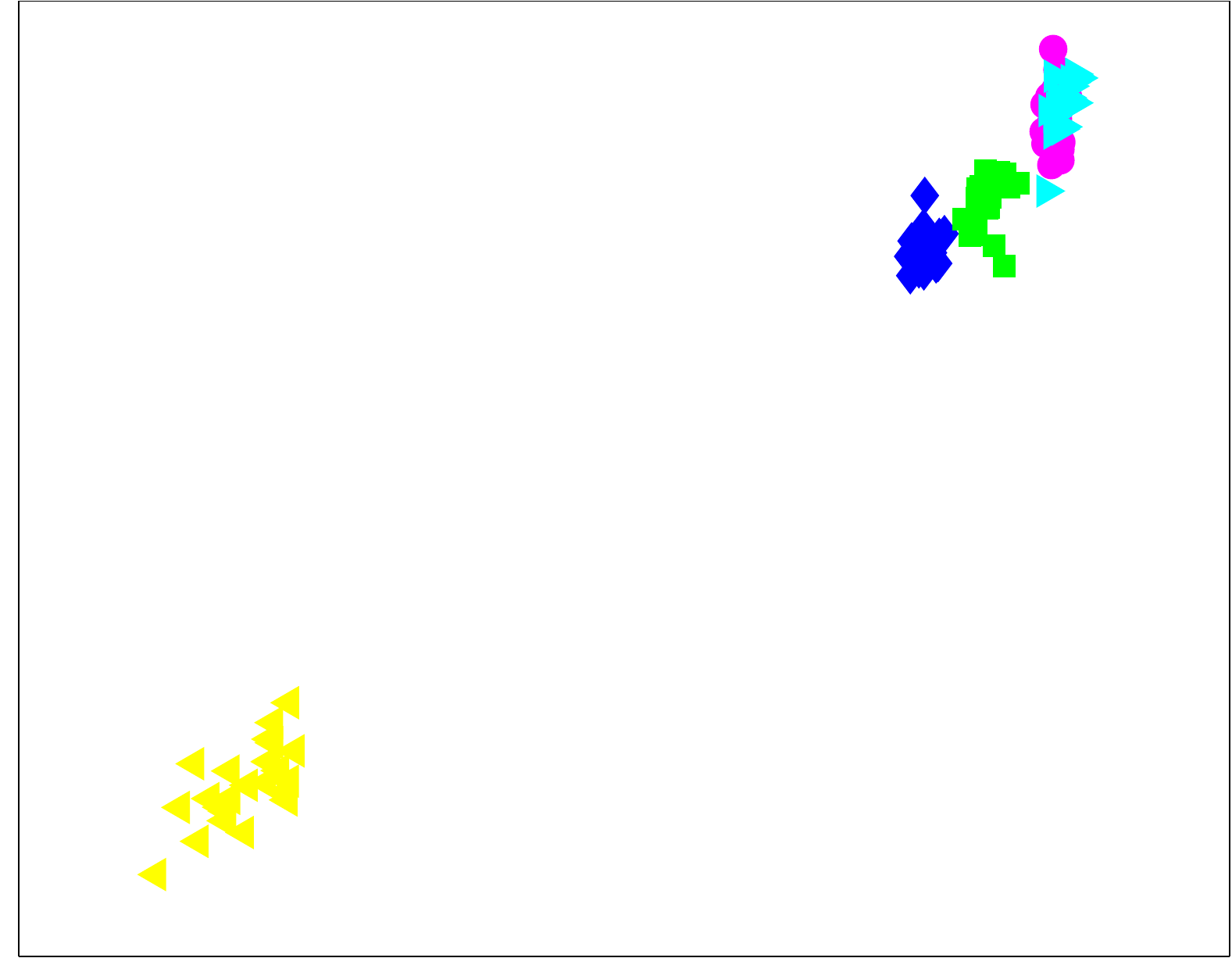}}\hspace{-0.2cm}
\subfigure[OLDA, Coil20]{
\centering \includegraphics[type=pdf,ext=.pdf,read=.pdf,width=0.24\textwidth,height=0.187\textwidth]{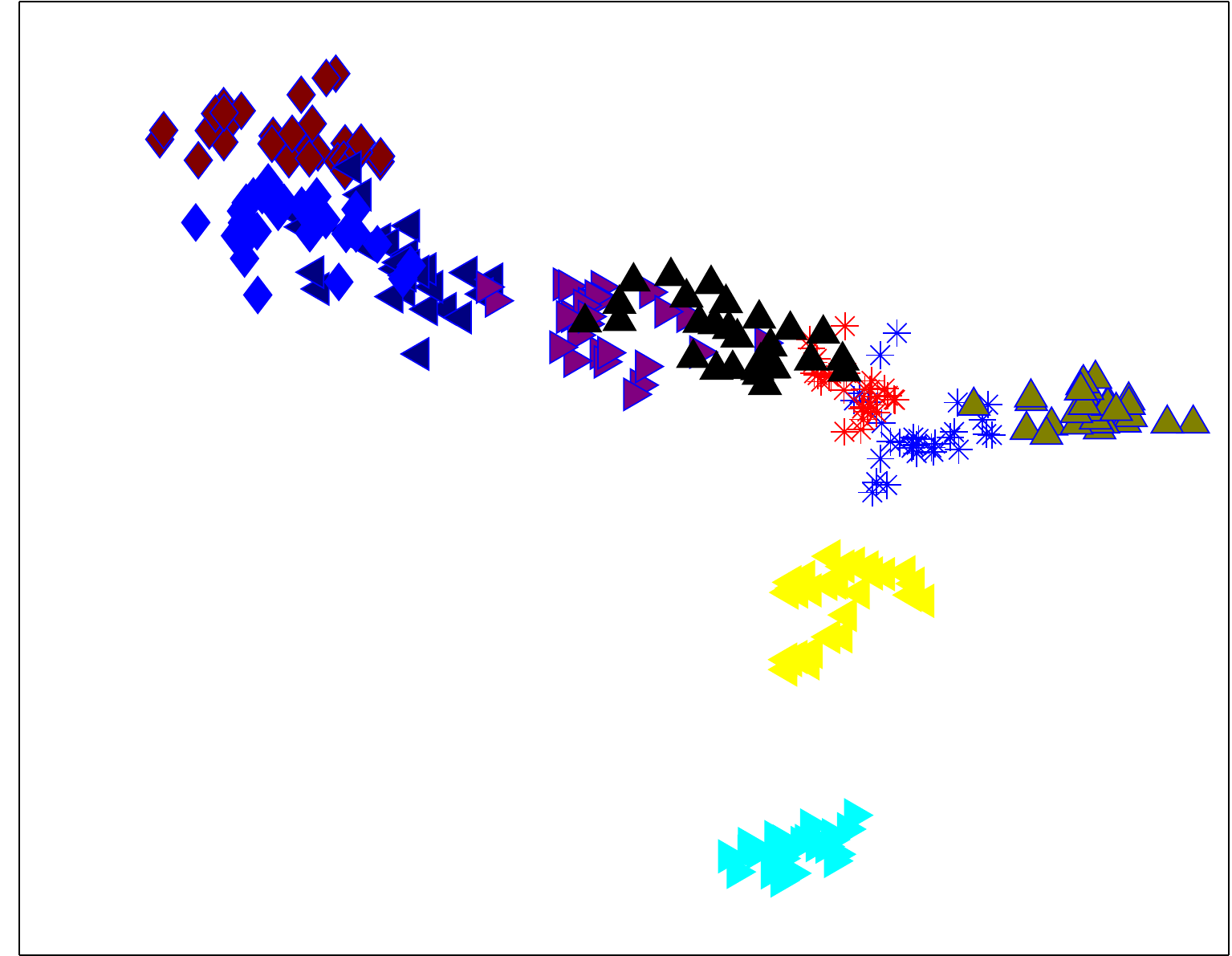}}\hspace{-0.2cm}
 \subfigure[OLDA, BA1]{
\centering \includegraphics[type=pdf,ext=.pdf,read=.pdf,width=0.24\textwidth,height=0.187\textwidth]{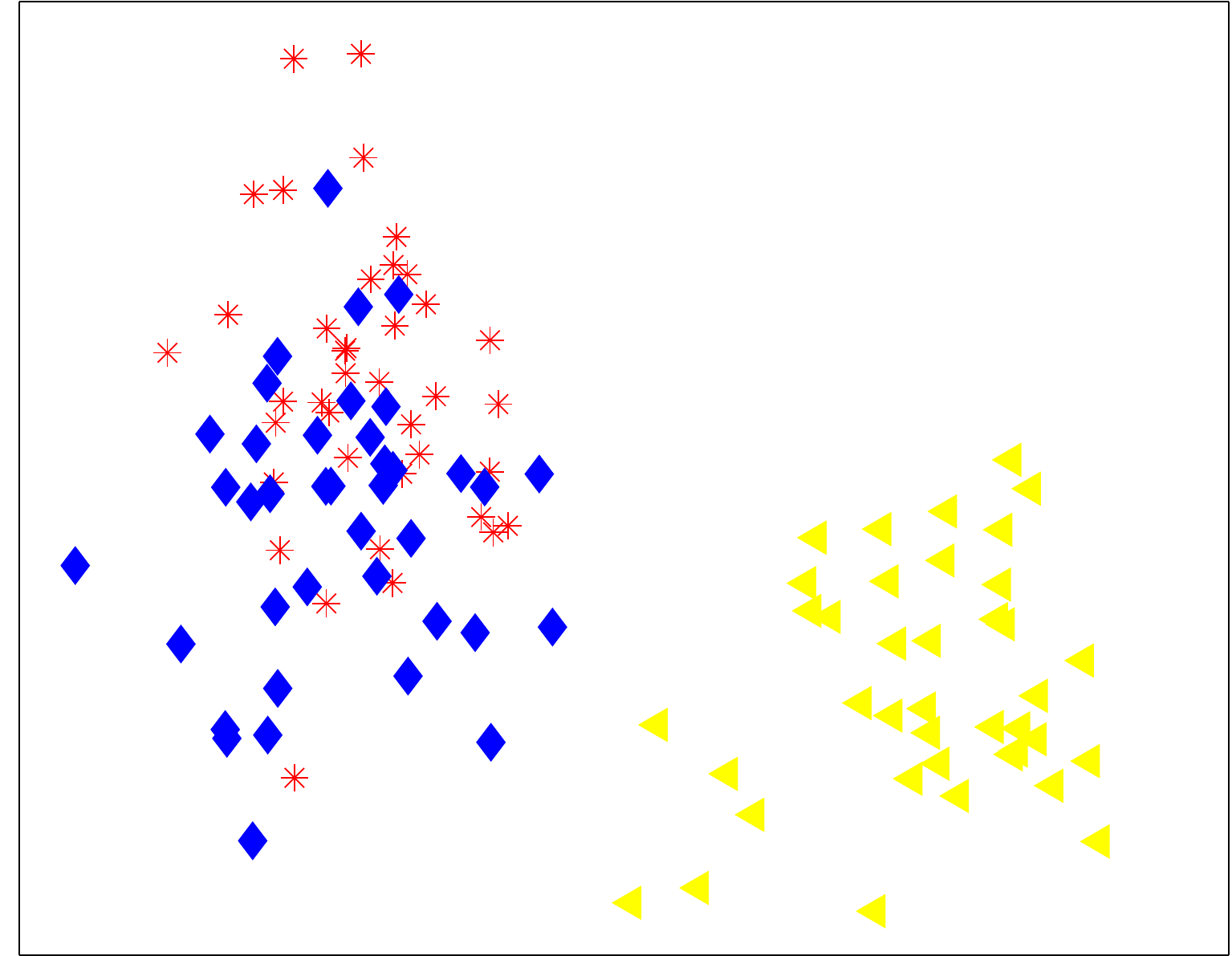}}
\\
 \subfigure[\wlda, Yale]{
\centering \includegraphics[type=pdf,ext=.pdf,read=.pdf,width=0.24\textwidth,height=0.194\textwidth]{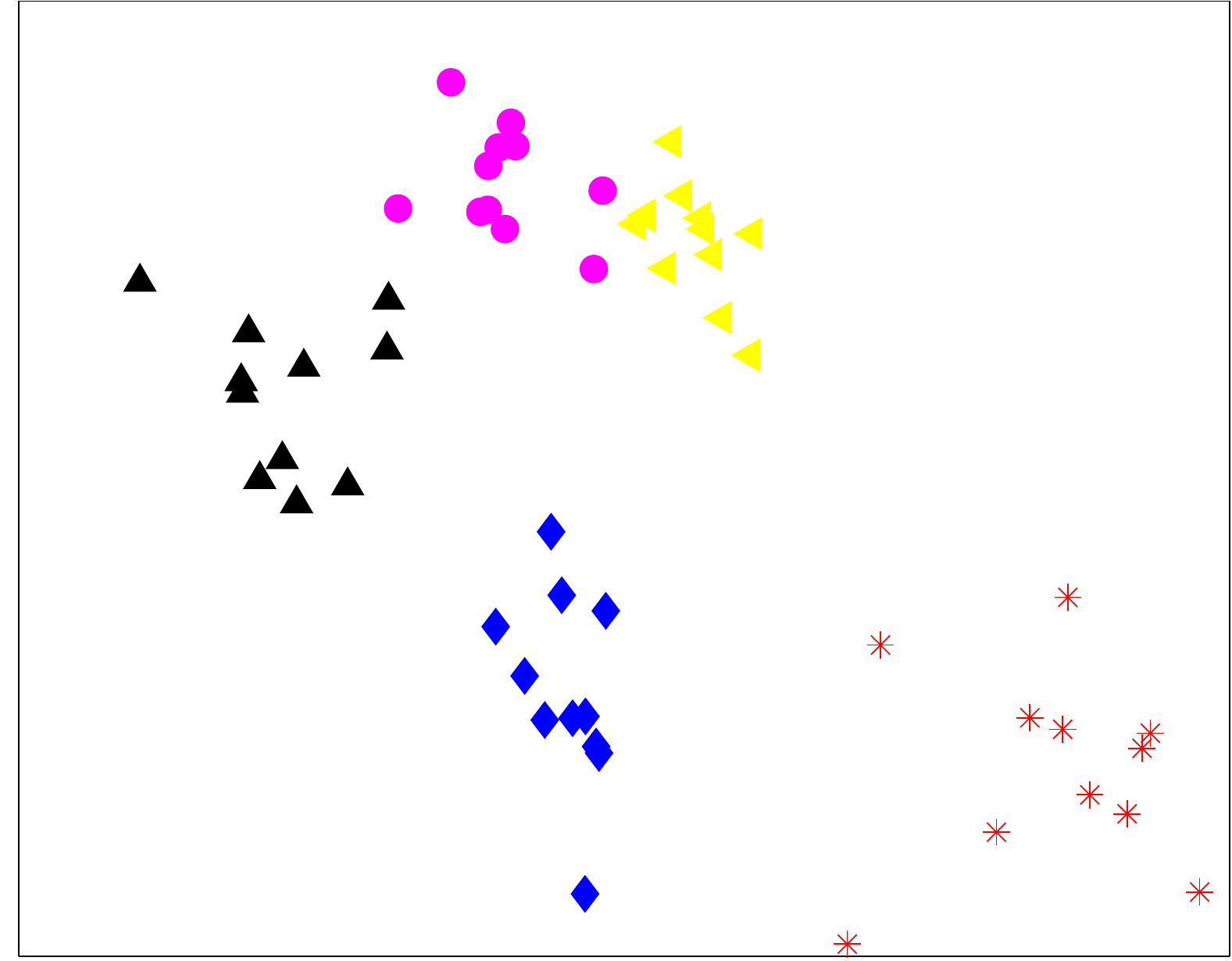}}\hspace{-0.2cm}
 \subfigure[\wlda, ALOI]{
\centering \includegraphics[type=pdf,ext=.pdf,read=.pdf,width=0.24\textwidth,height=0.195\textwidth]{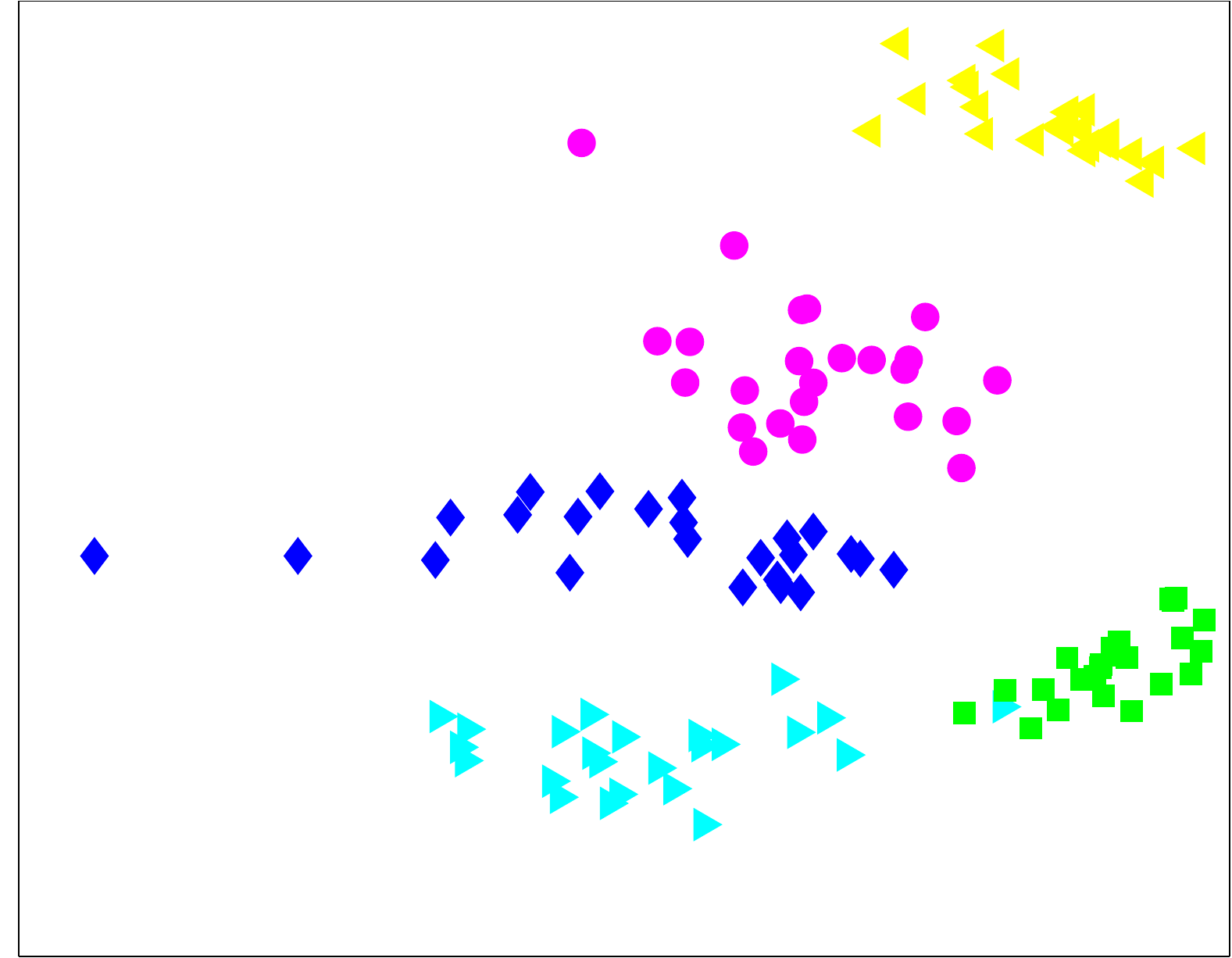}}\hspace{-0.2cm}
\subfigure[\wlda, Coil20]{
\centering \includegraphics[type=pdf,ext=.pdf,read=.pdf,width=0.24\textwidth,height=0.195\textwidth]{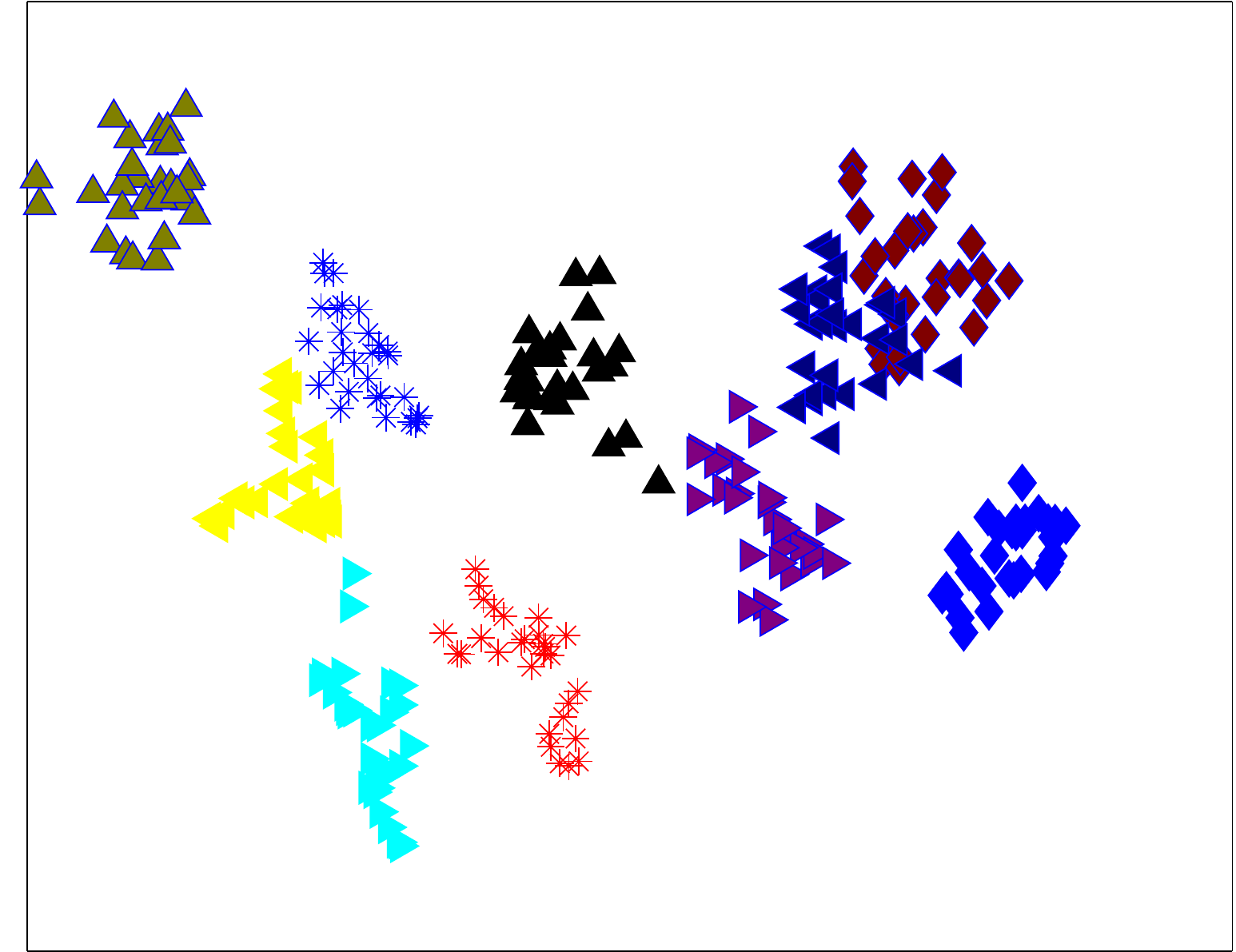}}\hspace{-0.2cm}
 \subfigure[\wlda, BA1]{
\centering \includegraphics[type=pdf,ext=.pdf,read=.pdf,width=0.24\textwidth,height=0.195\textwidth]{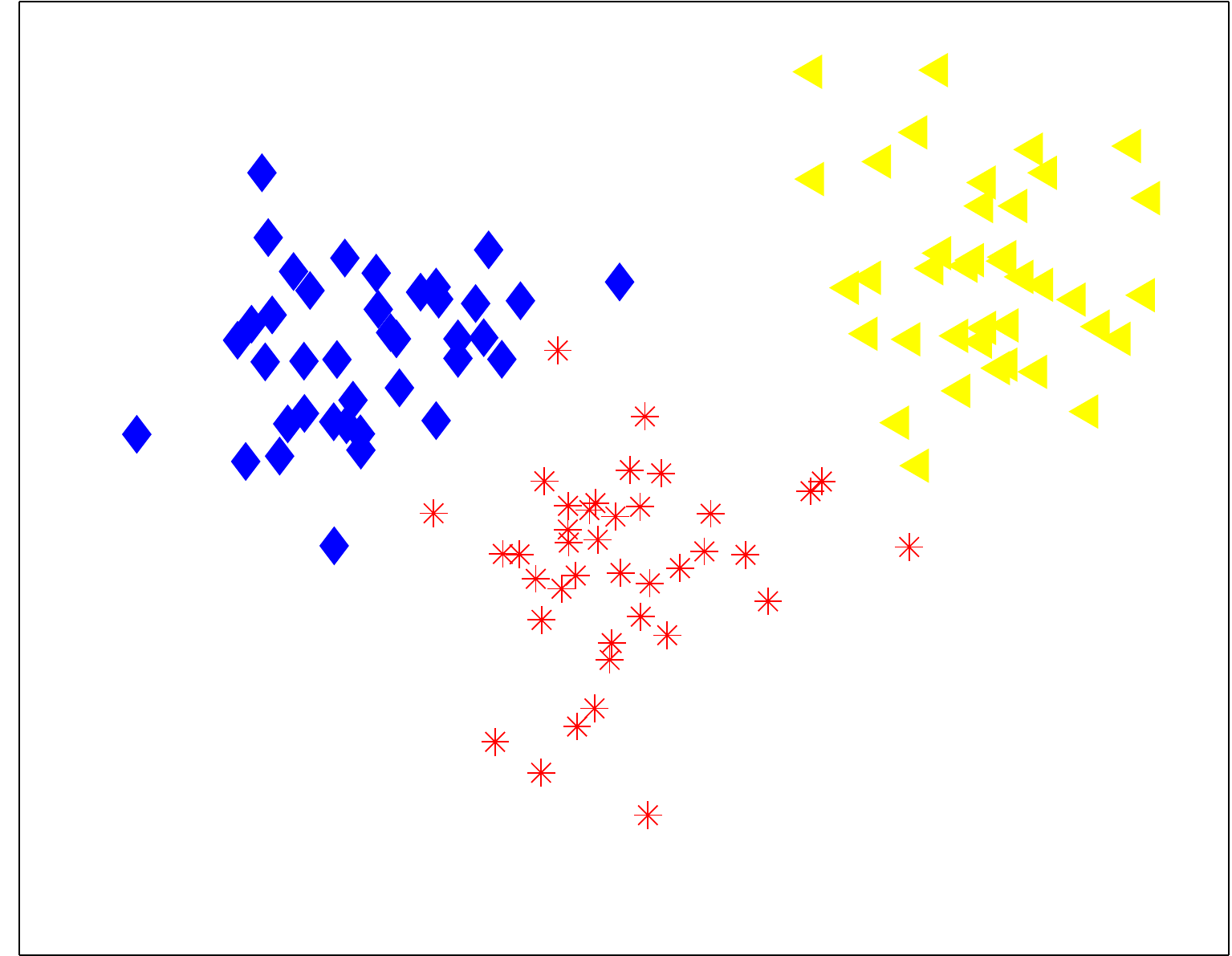}}

   \caption{Visualization of dimensionality reduction results of PCA, LDA, OLDA and \wlda \xspace applied to (from left) the Yale, ALOI, Coil20 and BA1 datasets. The feature dimensions are reduced from $50$, $100$, $50$ and $80$ (which have been processed by PCA in advance) respectively to $2$.
The figures show explicitly that the best classification results are achieved by \wlda.}
\label{fig:visual}

\end{figure*}

This section presents the experimental results of data visualization. The input data is projected to two dimensional subspace using the linear transformation matrix learned by PCA, LDA, OLDA and the proposed algorithm \wlda.
The data distributions after projection are shown in Fig.~\ref{fig:visual}. Several datasets are evaluated: Yale face dataset~\cite{Yale1997} with 5 classes ($5$th to $9$th) adopted, ALOI object dataset~\cite{Geusebroek20005} with 5 classes ($10$th to $14$th) used, Coil20 object dataset~\cite{coil20} with 10 classes ($1$st to $10$th) employed, and the Binary Alphadigits dateset~\cite{BA} with images of digits $3$, $4$ and $5$ used.
As shown in Fig.~\ref{fig:visual}, \wlda \xspace separates data better than PCA, LDA and OLDA on those datasets. PCA (unsupervised) preserves directions with the largest variance but much of the discriminant information is lost. LDA (supervised) considers the scatter measure in the average view, so the poor separations of two classes are probably to be concealed by other good separations. For example, in Fig.~\ref{fig:visual} (6), LDA separates most of classes, but fails to separate one pair. Because some classes are separated far from each other, the LDA criterion cannot demonstrate the fact that one pair of classes have not been separated yet. OLDA solved the nonsingularity limitation of scatter matrices in LDA, however, the scatter measures are still from the average viewpoint. Alternatively, \wlda \xspace tries to separate data from the worst-case viewpoint, so the separation of every class-pair is taken into account.

